\documentclass[11pt,a4paper]{article}
\pdfoutput=1

\usepackage{amsmath}
\usepackage{amssymb}
\usepackage{graphicx}
\usepackage{url}
\usepackage{pict2e}
\usepackage{pdfpages}
\usepackage{a4wide}

\usepackage[bookmarksopen,pdfstartview={FitH},pdfpagelabels=false,pdfborder={0 0 0}]{hyperref}
\def\reportno#1{\gdef\@reportno{#1}}

\newcommand{\wi}{\emph{wi}}
\newcommand{\ii}{\emph{ii}}
\newcommand{\well}{\wi-consultant}
\newcommand{\ill}{\ii-consultant}
\newcommand{\wells}{\wi-consultants}
\newcommand{\ills}{\ii-consultants}
\newcommand{\Arguments}{\mathit{Ar}}
\newcommand{\attack}{\mathit{att}}

\newcommand{\leftFig}[1]{
\hspace{-1.5cm}
 \begin{minipage}{0.5\textwidth}
  \centering
  \includegraphics[width=0.98\hsize]{RESULTS_corr/#1}\\
 \end{minipage}
\hfill}

\newcommand{\rightFig}[1]{
\hspace{-0.3cm}
\begin{minipage}{0.5\textwidth}
  \centering
  \includegraphics[width=0.98\hsize]{RESULTS_corr/#1}\\
\end{minipage}
}

\begin{document}

\includepdf[pages={1-2}]{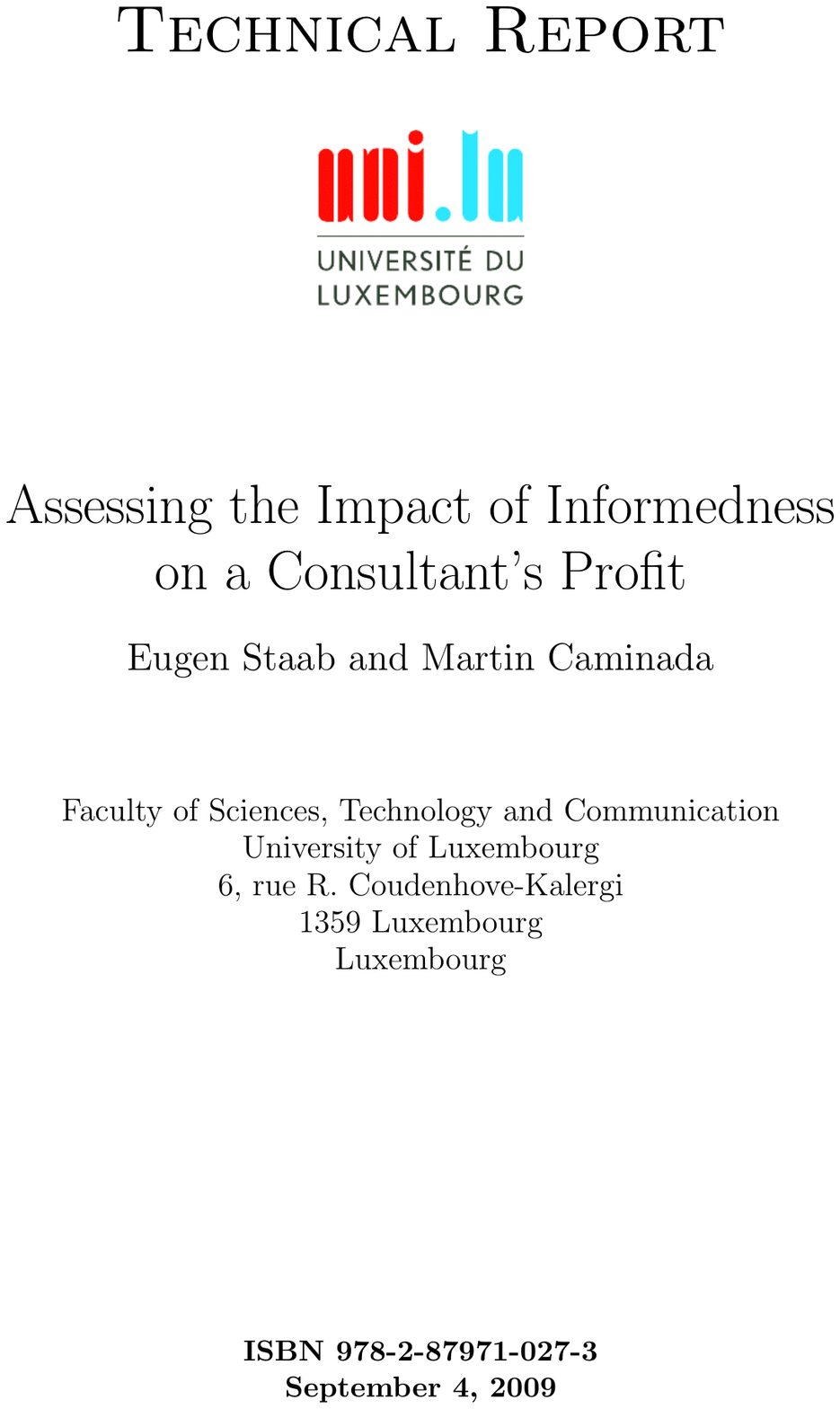}

\title{Assessing the Impact of Informedness on a Consultant's Profit}
\author{Eugen Staab and Martin Caminada \\ 
\small University of Luxembourg \\ 
\small\{eugen.staab,martin.caminada\}@uni.lu\\
}

\date{}

\maketitle

\begin{abstract}
We study the notion of informedness in a client-consultant setting. Using a software simulator, we examine the extent to which it pays off for consultants to provide their clients with advice that is well-informed, or with advice that is merely meant to appear to be well-informed. The latter strategy is beneficial in that it costs less resources to keep up-to-date, but carries the risk of a decreased reputation if the clients discover the low level of informedness of the consultant. Our experimental results indicate that under different circumstances, different strategies yield the optimal results (net profit) for the consultants.
\end{abstract}

\section{Introduction}
\label{sec:introduction}

When purchasing information, one wants to be sure of the quality of the
information in question. However, if one is not an expert oneself in the
relevant domain, assessing the quality of information can be difficult. 
For the sellers of information (which we will simply refer to
as ``the consultants'') this provides an incentive for dishonesty. After
all, gaining real expertise costs significant efforts as well as time and money. However, if the consumer of information (which we will refer to
as ``the client'') has difficulties assessing the quality of the provided
information, then why not pretend to have a higher level of expertise
than one actually has? As long as the chance that the client detects this
dishonesty is low, a consultant can charge the same price for his advice,
yet spend less resources on maintaining up-to-date of the state of the art.
Moreover, the more consultants decide to take it easy, the less likely it
is that the clients will find out about it. This is because once a critical
mass of consultants gives ill-informed advice, it becomes more and more
difficult for the clients to obtain the kind of well-informed advice that
they need in order to detect the ill-informedness of the other advice.
This implies that once a critical mass of low expertise consultants has
become established, the incentive for consultants to take it easy will
increase, since the chance of discovery has decreased.

The issue of low quality information has been studied in \cite{Fra05,Cam09a}.
What is new, however, is that we have now developed a software simulator
that is able to assess the pay-off for the consultants of either a strategy
of hard work or a strategy of taking it easy when it comes to staying up to
date with the state of the art. In particular, we are able to provide 
qualitative insight on which strategy yields the most profitable results
under which circumstances.

\section{Argumentation and Informedness}

The aim of this section is to examine how argumentation can play a role 
to examine the concept of informedness, which can be seen as background
theory for the remaining, more practical part of this paper.

In standard epistemic logic (S5), informedness is basically a 
binary phenomenon. One either has knowledge about a proposition $p$ or one does not.
It is, however, also possible to provide a more subtle account of the
extent to which one is informed about the validity of proposition $p$. 
Suppose Alex thinks that Hortis Bank is on the brink of bankruptcy because 
it has massively invested in mortgage backed securities. Bob also thinks 
that Hortis is on the brink of bankruptcy because of the mortgage backed 
securities. Bob has also read an interview in which the finance minister 
promises that the state will support Hortis if needed. However, Bob also 
knows that the liabilities of Hortis are so big that not even the state 
will be able to provide significant help to avert bankruptcy. From the 
perspective of formal argumentation \cite{Dun95}, Bob has three arguments 
at his disposal.\\
$A$: Hortis Bank is on the brink of bankruptcy, 
	because of the mortgage backed securities.\\
$B$: The state will save Hortis, because the finance minister promised so.\\
$C$: Not even the state has the financial means to save Hortis.\\
Here, argument $B$ attacks $A$, and argument $C$ attacks $B$ (see eq. 
\ref{eq:reinstatement}). In most approaches to formal argumentation,
arguments $A$ and $C$ would be accepted and argument $B$ would be rejected.
%\begin{figure}[h]
%\begin{center}
%$A\leftarrow B\leftarrow C$
%%\includegraphics[scale=0.75]{reinstatement.eps}
%\end{center}
%\caption{Argument $C$ attacks $B$, and argument $B$ attacks $A$.
%	\label{eq:reinstatement}}
%\end{figure}
\begin{align}
A\longleftarrow B\longleftarrow C
\label{eq:reinstatement}
\end{align}
Assume that Alex has only argument $A$ to his disposal. Then it seems to regard
Bob as more informed with respect to proposition $p$ (``Hortis Bank is on
the brink of bankruptcy'') since he has a better knowledge of the facts relevant
for this proposition and is also in a better position to defend it in the
face of criticism. 

The most feasible way to determine whether someone is informed on some given
issue is to evaluate whether he is up to date with the relevant arguments
and is able to defend his position in the face of criticism. One can say that 
agent $X$ is more informed than agent $Y$ if it has to its disposal a 
larger set of relevant arguments.

We will now provide a more formal account of how the concept of informedness 
could be described using formal argumentation. An \emph{argumentation
framework} \cite{Dun95} is a pair $(\Arguments, \attack)$ where $\Arguments$
is a set of arguments and $\attack$ is a binary relation on $\Arguments$.
An argumentation framework can be represented as a directed graph. For
instance, the argumentation framework $(\{A,B,C\}, \{(C,B),(B,A)\})$ is
represented in eq. \ref{eq:reinstatement}.

Arguments can be seen as defeasible derivations of a particular statement.
These defeasible derivations can then be attacked by statements of other 
defeasible derivations, hence the attack relationship. Given an argumentation
framework, an interesting question is what is the set (or sets) of arguments
that can collectively be accepted. Although this question has traditionally
been studied in terms of the various fixpoints of the characteristic function
\cite{Dun95}, it is equally well possible to use the approach of argument
labelings \cite{Cam06d,Cam07d,CW09a}. The idea is that each argument gets
exactly one label (accepted, rejected, or abstained), such that the result
satisfies the following constraints.
\begin{enumerate}
  \item If an argument is labeled accepted then all arguments that attack
	it must be labeled rejected.
  \item If an argument is labeled rejected then there must be at least
	one argument that attacks it and is labeled accepted.
  \item If an argument is labeled abstained then it must not be the case
	that all arguments that attack it are labeled rejected, and it
	must not be the case that there is an argument that attacks it and
	is labeled accepted.
\end{enumerate}
%A labeling is called admissible iff it satisfies at least the first two 
%constraints. 
A labeling is called complete iff it satisfies each of the above three 
constraints. As an example, the argumentation framework of eq.
\ref{eq:reinstatement} has exactly one complete labeling, in which
$A$ and $C$ are labeled accepted and $B$ is labeled rejected. In general,
an argumentation framework has one or more complete labelings.
Furthermore, the arguments labeled accepted in a complete labeling
form a complete extension in the sense of \cite{Dun95}. Other standard
argumentation concepts, like preferred, grounded and stable extensions 
can also be expressed in terms of labelings \cite{Cam06d}.

In essence, one can see a complete labeling as a reasonable position one
can take in the presence of the imperfect and conflicting information
expressed in the argumentation framework. An interesting question is
whether an argument \emph{can} be accepted (that is, whether the argument
is labeled accepted in at least one complete labeling) and whether an
argument \emph{has to be} accepted (that is, whether the argument is labeled
accepted in each complete labeling). These two questions can be answered
using formal discussion games \cite{PS97,VP00,Cam04b,CW09a}. For instance,
in the argumentation framework of eq. \ref{eq:reinstatement}, a possible
discussion would go as follows.\\
Proponent: Argument $A$ has to be accepted.\\
Opponent: But perhaps $A$'s attacker $B$ does not have to be rejected.\\
Proponent: $B$ has to be rejected because $B$'s attacker $C$ has to be accepted.\\
The precise rules which such discussions have to follow are described in
\cite{PS97,VP00,Cam04b,CW09a}. We say that argument $A$ can be \emph{defended}
iff the proponent has a winning strategy for $A$. We say that argument $A$
can be \emph{denied} iff the opponent has a winning strategy against $A$.

If informedness is defined as justified belief, and justified is being 
interpreted as defensible in a rational discussion, then formal discussion 
games can serve as a way to examine whether an agent is informed with respect 
to proposition $p$, even in cases where one cannot directly determine the truth 
or falsity of $p$ in the objective world. An agent is informed on $p$ iff it 
has an argument for $p$ that it is able to defend in the face of criticism.

The dialectical approach to knowledge also allows for the distinction of
various grades of informedness. That is, an agent $X$ can be perceived to be
at least as informed as agent $Y$ w.r.t. argument $A$
iff either $X$ and $Y$ originally disagreed on the status of $A$ but combining
their information the position of $X$ is confirmed, or $X$ and $Y$ originally
agreed on the status of $A$ and in every case where $Y$ is able to maintain its
position in the presence of criticism from agent $Z$, $X$ is also able to
maintain its position in the presence of the same criticism.

When $AF_1 = (\Arguments_1, \attack_1)$ and $AF_2 = (\Arguments_2, \attack_2)$
are argumentation frameworks, we write $AF_1 \sqcup AF_2$ as a shorthand
for $(\Arguments_1 \cup \Arguments_2, \attack_1 \cup \attack_2)$, and
$AF_1 \sqsubseteq AF_2$ as a shorthand for $\Arguments_1 \subseteq \Arguments_2
\wedge \attack_1 \subseteq \attack_2$. Formally, agent $X$ is at least as
knowledgeable about argument $A$ as agent $Y$ iff:
\begin{enumerate}
  \item $A$ can be defended using $AF_X$ (that is, if $X$ assumes the role
	of the proponent of $A$ then it has a winning strategy using the 
	argumentation framework of $X$), $A$ can be denied using $AF_Y$
	(that is, if $Y$ assumes the role of the opponent than it has a
	winning strategy using the argumentation framework of $Y$), 
	but $A$ can be defended using $AF_X \sqcup AF_Y$, or
  \item $A$ can be denied using $AF_X$, $A$ can be defended using $AF_Y$, 
	but $A$ can be denied $AF_X \sqcup AF_Y$, or
  \item $A$ can be defended using $AF_X$ and can be defended using $AF_Y$,
	and for each $AF_Z$ such that $A$ can be defended using $AF_Y \sqcup
	AF_Z$ it holds that $A$ can also be defended using $AF_X \sqcup AF_Z$,
  \item $A$ can be denied using $AF_X$ and can be denied using $AF_Y$,
	and for each $AF_Z$ such that $A$ can be denied using $AF_Y \sqcup
	AF_Z$ it holds that $A$ can be denied using $AF_X \sqcup AF_Z$.
\end{enumerate}
Naturally, it follows that if $AF_Y \sqsubseteq AF_X$ then $X$ is at least
as informed w.r.t. each argument in $AF_Y$ as $Y$.

In the example mentioned earlier (eq. \ref{eq:reinstatement}) Alex
has access only to argument $A$, and Bob has access to arguments $A$, $B$
and $C$. Suppose a third person (Charles) has access only to arguments $A$
and $B$. Then we say that Bob is more informed than Alex w.r.t. argument 
$A$ because Bob can maintain his position on $A$ (accepted) while facing 
criticism from Charles, where Alex cannot. $A$ more controversial consequence 
is that Charles is also more informed than Alex w.r.t. argument $A$, 
even though from the global perspective, Charles has the ``wrong'' position 
on argument $A$ (rejected instead of accepted). This is compensated by the 
fact that Bob, in his turn, is more informed than Charles w.r.t. 
argument $A$. As an analogy, it would be fair to consider Newton as more 
informed than his predecessors, even though his work has later been 
attacked by more advanced theories.

\section{Simulation}
We developed a software-simulator in order to better understand the impact of a consultant's informedness on his profit.
The simulator has the objective to reveal when it pays off for a consultant to be \emph{ill-informed}, i.e. less well-informed as is the state of the art. 
If ill-informedness is more profitable for a consultant than being well-informed, the spread of outdated and possibly wrong facts or arguments is preprogrammed.
Of course, such a situation would be harmful to our society, and so a study of its causes seems to be important.
%In this section, we describe the simulator in detail, show some results and give an interpretation of these results.

In the following, we describe the simulator (Sect. \ref{sec:simulator}), show results (Sect. \ref{sec:results}) and discuss these (Sect. \ref{sec:analysis}).

\subsection{The Simulator}
\label{sec:simulator}
In the first part of this section, we describe the client-consultant scenario that we aim to simulate.
We then detail the considered argumentation framework. %on which the consultants try to advise their clients.
Two different strategies of the consultants concerning the acquisition of newly available information are defined afterwards, one representing well-informed consultants and the other representing ill-informed consultants. 
Finally, we specify the procedure of how clients select their consultants.

\subsubsection{Client-Consultant Scenario}
The scenario is basically defined by a set of clients, a set of consultants, and a finite number of rounds in each of which the clients seek advice from the consultants. 
In each round, consultants acquire new information, either from other sources (researchers, analysts, etc.), or do investigations on their own. 
We model this acquisition of information simply as fixed costs per piece of information; in our case, this information comes in the form of arguments as we will see later. 
%The consultants use the acquired information to advise their clients. 
The more clients request a certain consultant's advice, the lower his price for a consultation can be. 
Note that this actually explains why the intermediary role of consultants exists. 

Clients are free to choose their consultant, and so select in each round the consultant that they think is currently the most appropriate one. 
In our simulations, this selection is based on price and reputation of the consultants as it will be detailed later.

\subsubsection{Argumentation Framework}
We consider the following argumentation structure consisting of $N_\text{arg}$ many arguments:
\begin{align}
A_1 \longleftarrow A_2 \longleftarrow \dots \longleftarrow A_{N_\text{arg}}\;.
\end{align}
Here, any argument $A_i$ (for $1<i\leq N_\text{arg}$) defeats its predecessor argument $A_{i-1}$.
If $N_\text{arg}$ is even, then all arguments $A_i$ where $i$ is even, are ``in'', and all other arguments are ``out''. 
If $N_\text{arg}$ is odd, it is the other way around.

At the beginning of a simulation, only argument $A_1$ is known to the consultants and only this argument is known in the whole society, i.e. it represents the ``state of the art''. 
To model the discovery/emergence of new information, we make a certain number of new arguments available to the consultants in each round.
This represents the evolution of the state of the art.
The number of new arguments per round will be fixed for a simulation and is denoted by $\Delta N_\text{arg}$. 
We assume that the consultants extend their already known chain of arguments with new arguments always in a seamless manner, i.e. without gaps.
This assumption was made in order to be in line with argument games (such as described in \cite{PS97,VP00,Cam04b})
where each uttered argument is a reaction to a previously uttered argument, thus satisfying the property of
\emph{relevance} \cite{Cam06c}.

For a better understanding, the following shows the structure of the chain of arguments at any round of the simulation ($k\leq i$ must hold):
\begin{align}
\overbrace{\underbrace{A_1 \longleftarrow \dots \longleftarrow A_k}_{\text{known to certain consult.}} \longleftarrow \dots \longleftarrow A_i}^{\text{``state of the art''}} \longleftarrow \overbrace{A_{i+1} \longleftarrow  \dots \longleftarrow  A_{i+\Delta N_\text{arg}}}^{\text{becoming available next round}} \longleftarrow \dots \longleftarrow A_{N_\text{arg}}\;.
\end{align}

Consultants generally want to provide the least amount of information needed for a consultation, because this way they can give more consultations. 
However, consultants generally want to give good advice at the same time, to increase their reputation.
We assume that consultants believe that the latest argument they know is the most justified one, because it is closest to the current state of the art.
As a consequence, in order to give good advice, they only consult arguments that are compliant to the latest argument they possess, i.e. arguments whose index has the same parity as the latest argument they know. 
To provide the least amount of information at the same time, a rational consultant acts as follows: he provides the client with \emph{two} arguments, if the latest argument known to the client is of the same parity as the latest argument known to the consultant, and with \emph{one} argument otherwise.
The latest argument known to the client is updated accordingly.

To keep things simple, the cost of an argument is set to a constant $c_\text{arg}$. 
So, to get the knowledge about argument $A_{10}$, the consultant has an overall expense of $10 \cdot c_\text{arg}$ (recall that arguments can only be acquired in a row).
We write $n_\text{arg}$ to denote the total number of arguments acquired by a specific consultant (where $n_\text{arg}\leq N_\text{arg}$).
Thus, we model the expenses $E$ of a consultant as: 
\begin{align}
\label{eq:e}
E=n_{\text{arg}} \cdot c_{\text{arg}}\;.
\end{align}

\subsubsection{Strategies of the Consultants}
%todo: what is their their profit
We consider two types of consultants that deal differently with newly available arguments:
\begin{description}
	\item[well-informed (\wi):] \wells\ buy arguments as soon as these become available, since they want to be always up-to-date.
	\item[ill-informed (\ii):] \ills\ only buy arguments as to appear knowledgeable to the clients. That is, as soon as they notice that a client is as informed as they are, or even better informed, they buy a number of new arguments such that they know one argument more than this client.
\end{description}
Clearly, \ills\ can offer their consultations at a lower price. 
However, the reputation of a consultant decreases with each consultation where the client turns out to be as informed as the consultant -- and this happens more often to \ills. 

The turnover of a consultant is defined by the sum of prices the consultant was paid. 
Let $S$ be the multiset enumerating all prices paid by clients up to a certain round. 
So, $S$ represents the consultations where the consultant actually was better informed than the client, and thus was paid.
Then the turnover $T$ up to this point is defined as: 
\begin{align}
\label{eq:T}
T = \sum_{p\in S}{p}\;.
\end{align}

Finally, the profit $P$ of a consultant up to a certain round is the difference between his turnover and his expenses so far: 
\begin{align}
P=T-E\;. %\frac{\text{price}}{\text{arg}}
\end{align}

\subsubsection{Selection of a Consultant}
In our simulations, clients rate consultants according to two criteria: the consultant $i$'s current reputation $r_i$ and price $p_i$.
The two criteria are explained in more detail later in this section. 
For now, it suffices to know that they are normalized to $[0,1]$. 
To make both parameters ``positive'', the price $p_i$ will be expressed in the form of \emph{cheapness} $c_i$, i.e., $c_i=1-p_i$. 
This way, both a \emph{high} cheapness and a \emph{high} reputation characterize a good consultant.
A parameter $\alpha\in[0,1]$ defines which of the two criteria a client thinks is more important.
Consequently, a client chooses consultant $i$ with a probability proportional to:
\begin{align}
P_i:=\alpha\cdot c_i+(1-\alpha)\cdot r_i\;.
\end{align}
A high $\alpha$ favors the choice of cheaper consultants, while a low $\alpha$ favors the choice of more reputable consultants.

\paragraph{Price}
Let $\delta$ be the \emph{profit margin} of a consultant, with $\delta\in [0,\infty)$; then $\delta=0.5$ for example represents a typical profit margin of $50\%$. 
Using a certain profit margin, a consultant $i$ computes his current price ${p'}_i$:
\begin{align}
\label{price_model}
{p''}_i=(1+\delta)\frac{E}{|S|}\;.
\end{align}
Here, $\frac{E}{|S|}$ is a simple estimate to provide cost recovery, where $E$ models the expenses (see eq. \ref{eq:e}), and $|S|$ is the number of successful consultations so far (see eq. \ref{eq:T}). Still, no client would choose a consultant that is more expensive than the acquisition of the information itself. Hence, we limit the price to the cost of the two arguments that a consultant advises at most ($2\cdot c_\text{arg}$).
%\begin{align}
%{p'}_i=\min\{{p''}_i,2\cdot c_\text{arg}\}\;.
%\end{align}

Finally, we normalize the prices of all consultants to $[0,1]$:
\begin{align}
p_i = \frac{{p'}_i-\min_j{({p'}_j)}}{\max_j{({p'}_j)}-\min_j{({p'}_j})} \;.
\end{align}

%In short, the price of a consultant reflects the costs of the acquired information, his profit margin and the estimated number of clients requesting his advice.
%The profit margin is represented in percent, i.e. by a real number in $[0,\infty)$; e.g., $0.05$ stands for a profit margin of $5\%$.
%
%We let a consultant estimate the number of requesting clients for the current round as the simple moving average of the number of requesting clients from the last $r$ rounds.
%
%Given the expenses $E_i$, the turnover $T_i$ and the consultants profit margin $m$, then the amount $\delta$ of expenses not covered so far (in respect to the profit margin) is:
%\begin{align}
%\delta=\frac{T_i}{1+m}-E_i
%\end{align}
%In each round a consultant determines his current price as follows:
%\begin{align}
%p=m\cdot \frac{E_i}{}
%\end{align}

\paragraph{Reputation}
In our simulator, clients use a \emph{reputation system}~\cite{resnick00reputation} to share their experience about consultants.
This allows clients to better estimate the trustworthiness of the consultants and thus to better select their future consultants. 
We assume perfect conditions for the reputation system, since this will make it even harder for \ills\ to hold their ground. 
These perfect conditions consist of:
\begin{itemize}
	\item \emph{honest reporting} of the clients,
	\item all clients have the same idea of how to fuse the experiences with consultants, and so a global reputation score can be computed, and
	\item \emph{total information sharing}, i.e., every client shares his experience with every other client.	
\end{itemize}

The reputation system should be as simple as possible, and the only requirement we have is that it measures a consultant's performance relatively to the performance of other consultants.
We consider two different reputation systems. A very basic system (R1), and a more sophisticated one (R2).

\subparagraph{Reputation System R1}
In the reputation system R1, a client is satisfied with a consultation if the consultant is better informed than the client. 
This is motivated by the fact that clients cannot question the correctness of the consultants arguments since they are less well informed.
So, any consultation where the consultant is better informed than the client counts as a positive experience with that consultant. 
Consultations where this is not the case count as negative experiences. 
The clients share their experience and maintain for each consultant $i$ a global counter $n^+_i$ of positive experiences, and a global counter $n^-_i$ of negative experiences. 
Then an intermediate reputation score is computed as follows:
\begin{align}
{r'}_i=n^+_i-n^-_i \;.
\end{align}
To bring these scores into $[0,1]$, we normalize the intermediate values of all consultants and get the final global reputation score $r_i$:
\begin{align}
r_i = \frac{{r'}_i-\min_j{({r'}_j)}}{\max_j{({r'}_j)}-\min_j{({r'}_j})} \;.
\end{align}
This basic reputation score has the desired property: the overall performance of a consultant is measured relatively to the performance of other consultants. 
%To make the score more realistic, one could weight negative experiences stronger. 
%However, this would also make it only harder for \ills\ .

\subparagraph{Reputation System R2}
In this reputation system, a client reactively decides about positive and negative experiences.
Each client assumes that the latest argument he was advised on by a consultant is correct.
Then for all consultants $i$ that advised him with a conflicting argument in the past (i.e., where the parity of the argument's index is different), each client increases the global value $n^-_i$, and for all consultants that advised a compatible argument, $n^+_i$ is increased.
As in R1, if the consultant is not more informed than the client, $n^-_i$ is additionally increased.
Given $n^+_i$ and $n^-_i$, the computation of $r_i$ is the same as in R1.

%A reputation system . Though few producers or
%consumers of the ratings know one another, ,
%encourage trustworthy behavior, and deter
%participation by those who are unskilled
%or dishonest.

\subsection{Results}
\label{sec:results}

On the following pages we show simulation results to compare the profit of \wells\ and \ills\ for different parameters. 
We conducted the experiments by computing the mean and standard deviation of the consultants' profit over $2^8$ simulation runs. 
Each simulation consisted of $2^7$ rounds. 
We used $2^{10}$ clients and $2^7$ consultants. 

We varied the following parameters:
\begin{itemize}
  \item the number of arguments becoming available each round ($\Delta N_\text{arg}\in\{2,3,4(,5)\}$),
	\item the fraction of \ills\ ($f_\ii\in\{0.1,0.5,0.9\}$), 
	\item the profit margin ($\delta\in\{0.1,0.5\}$), and 
	\item the factor $\alpha$ that regulates the importance of price and reputation in the clients' consultant selection process ($\alpha\in[0,1]$, shown on the x-axis).
\end{itemize}
The figures on the left depict simulations with $\delta=0.1$, those on the right depict simulations with $\delta=0.5$.
Furthermore, we conducted all experiments for the two different reputation systems R1 and R2. 
At the end of this section, we provide an analysis of the shown results.\\

\noindent(Note: For making it easier to compare the different figures, we start on the next page.)

\vfill
\pagebreak
\subsubsection{Using R1}
\begin{itemize}
 
 \item Information rate $\Delta N_\text{arg}=2$:

 \begin{itemize}
	\item Fraction of \ills\ $f_{\ii}=0.1$:

	\leftFig{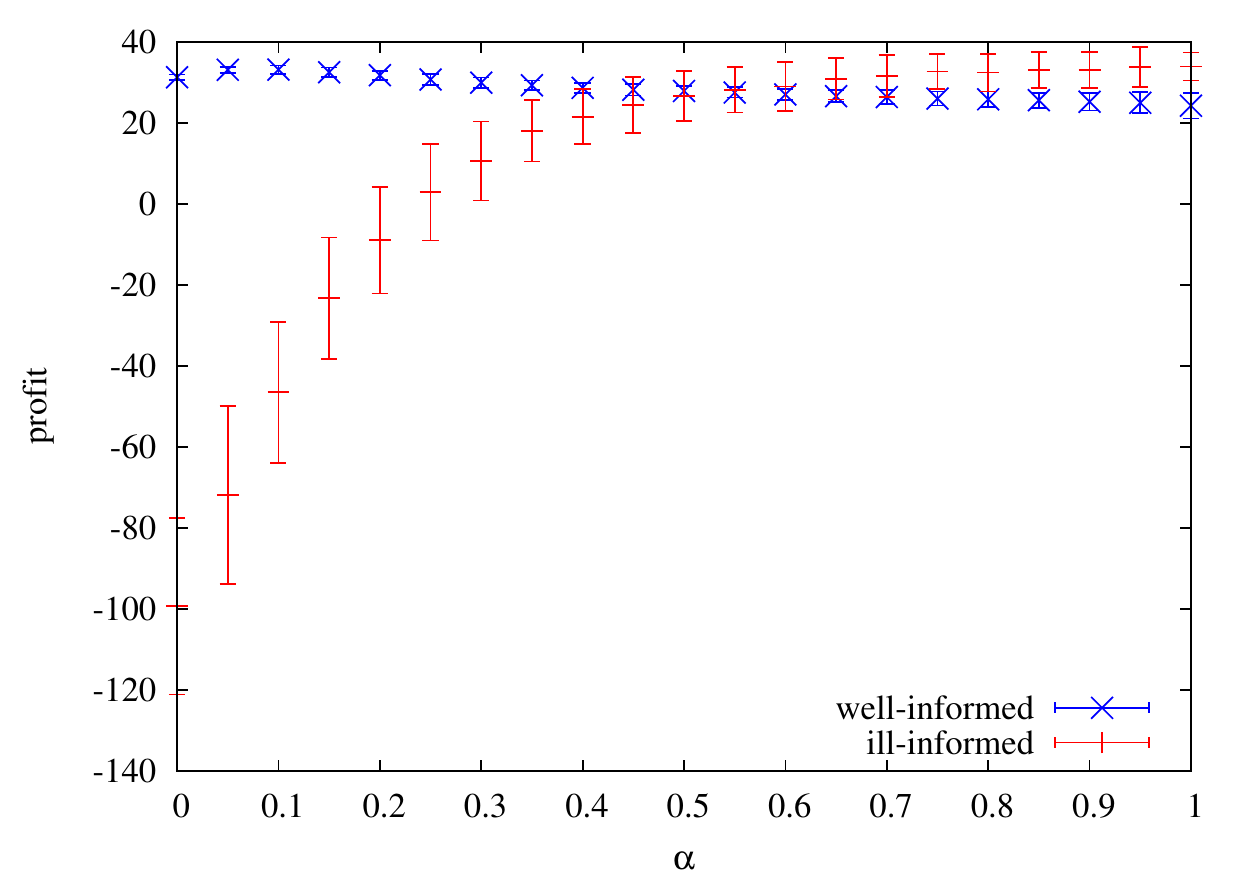}
	\rightFig{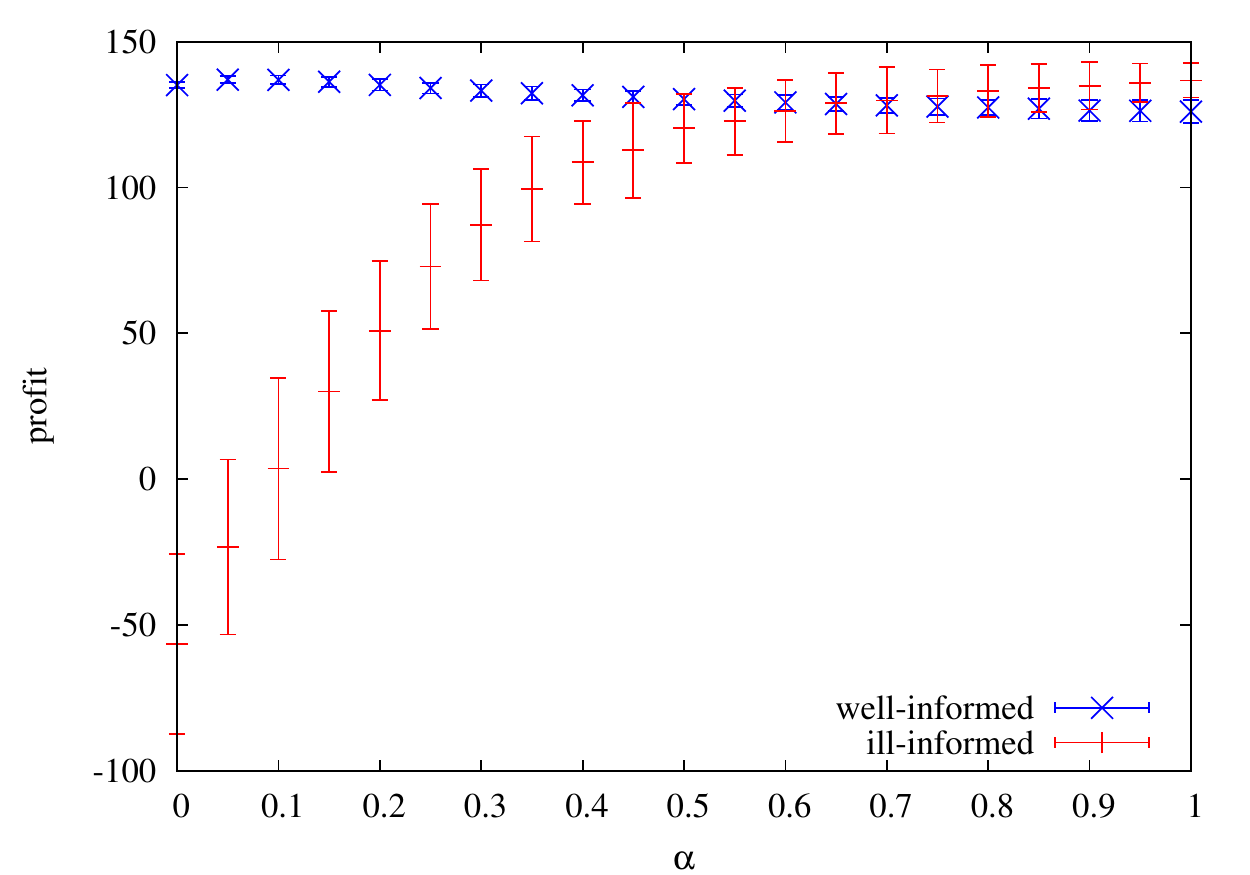}

	\item Fraction of \ills\ $f_{\ii}=0.5$
	
	\leftFig{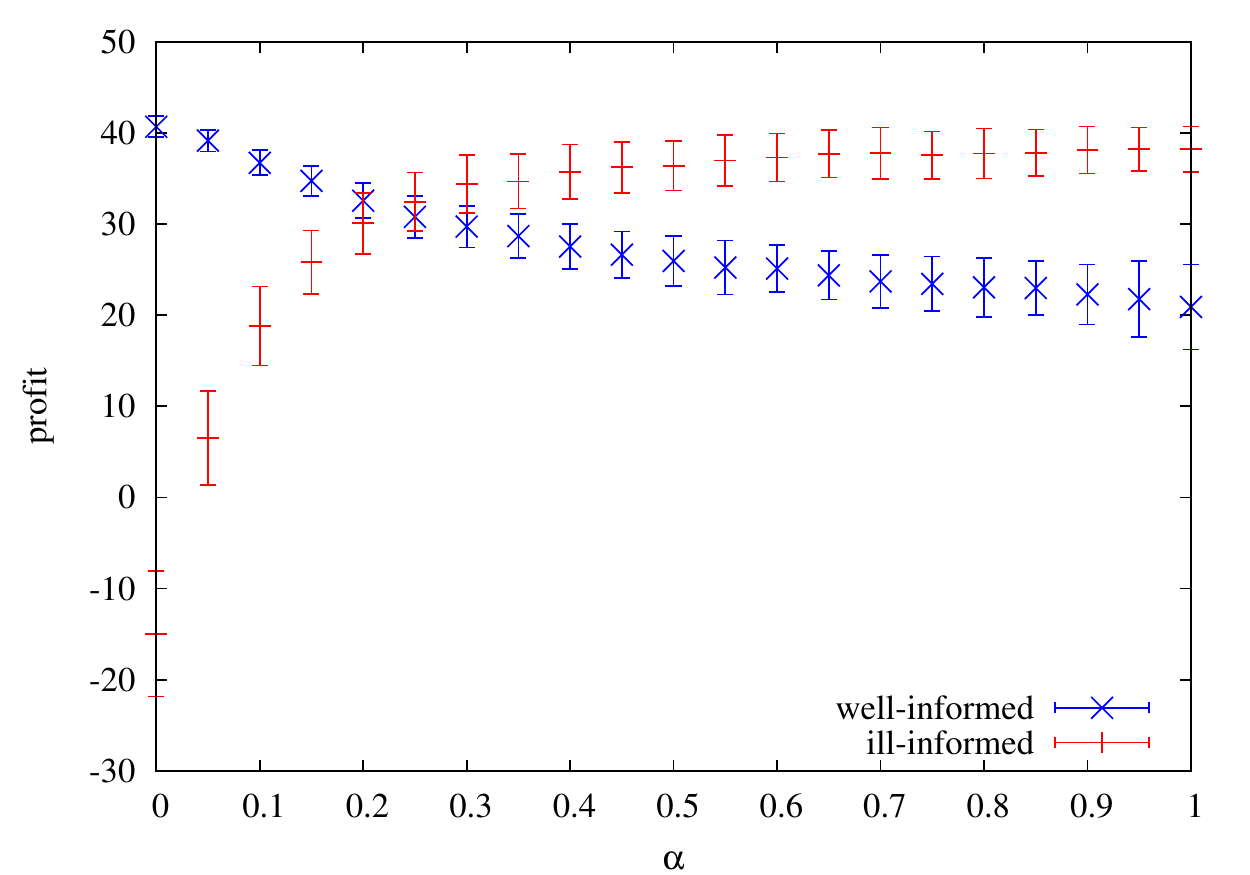}
	\rightFig{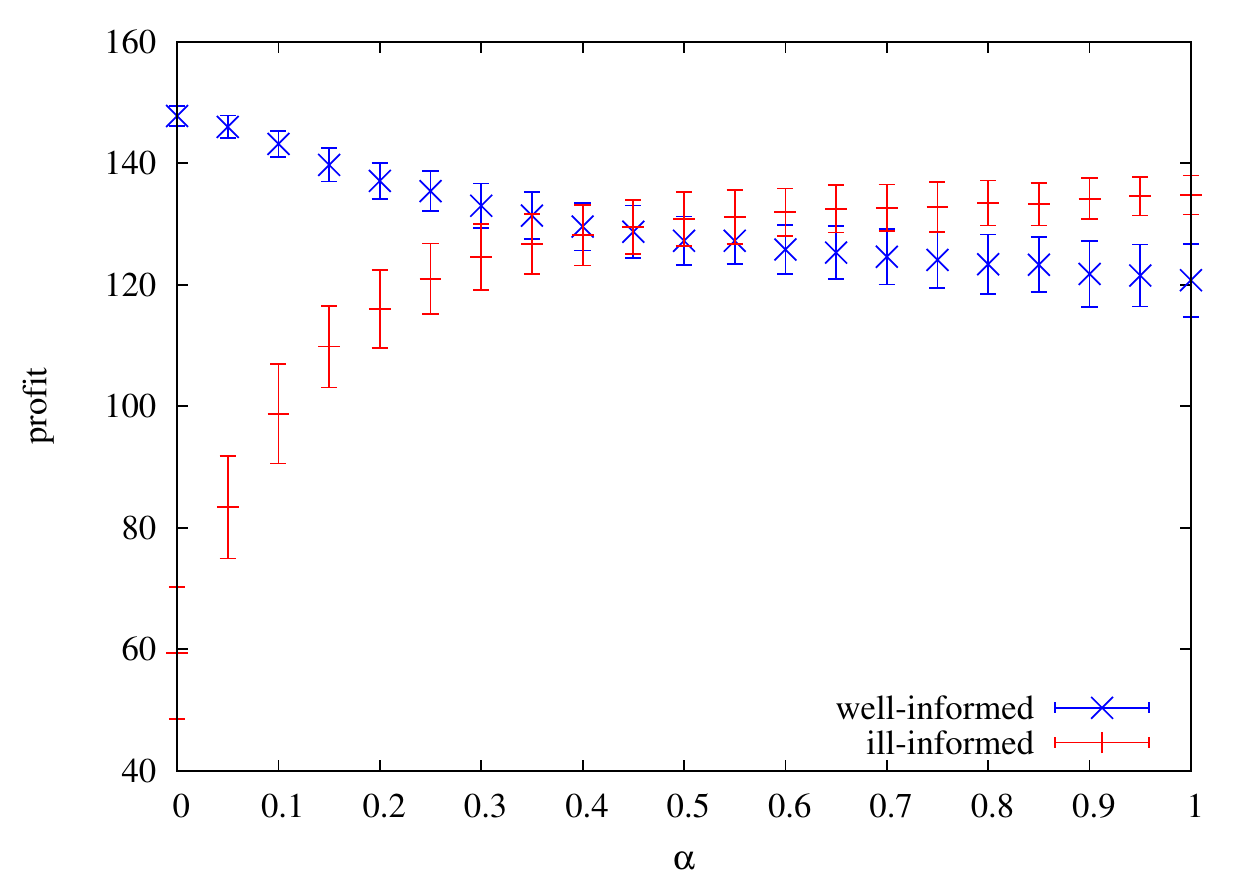}

	\item Fraction of \ills\ $f_{\ii}=0.9$

	\leftFig{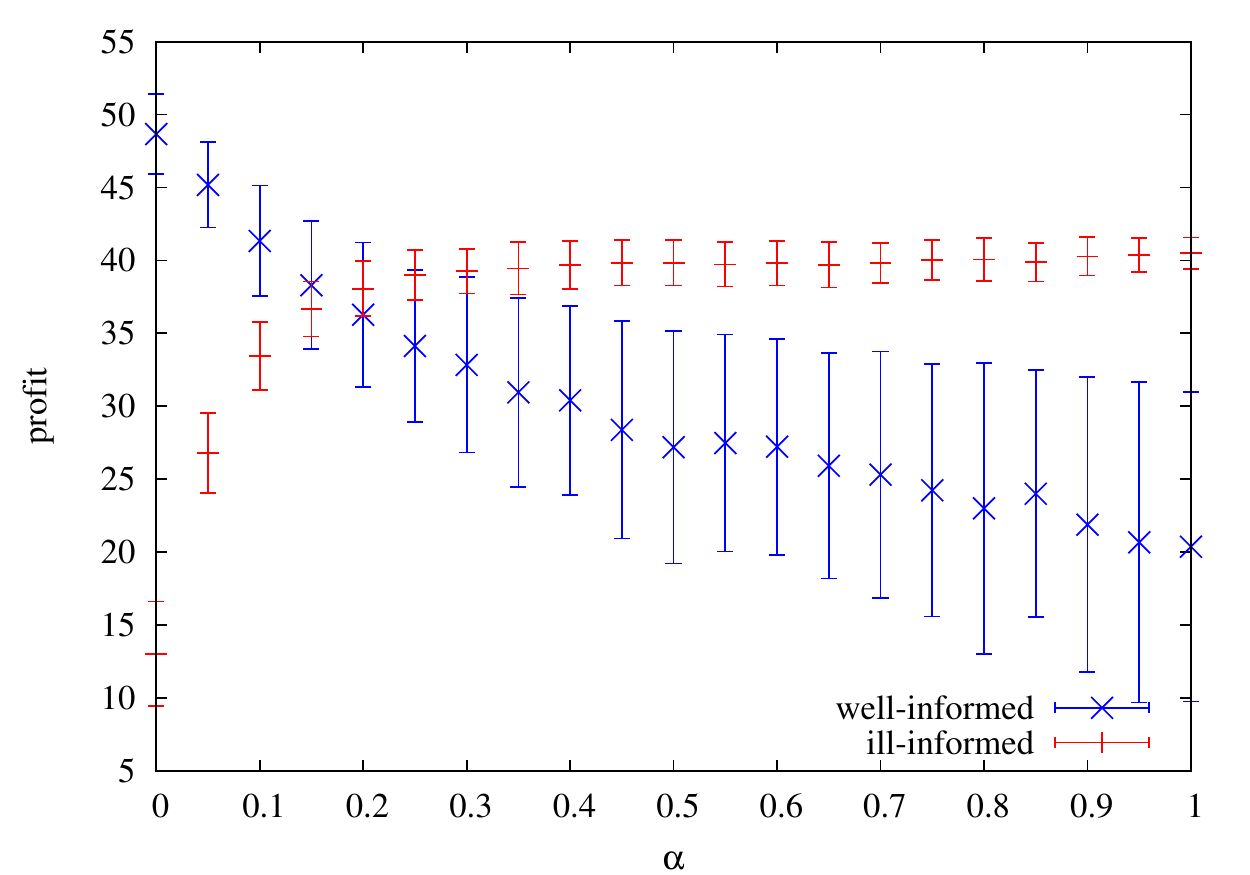}
	\rightFig{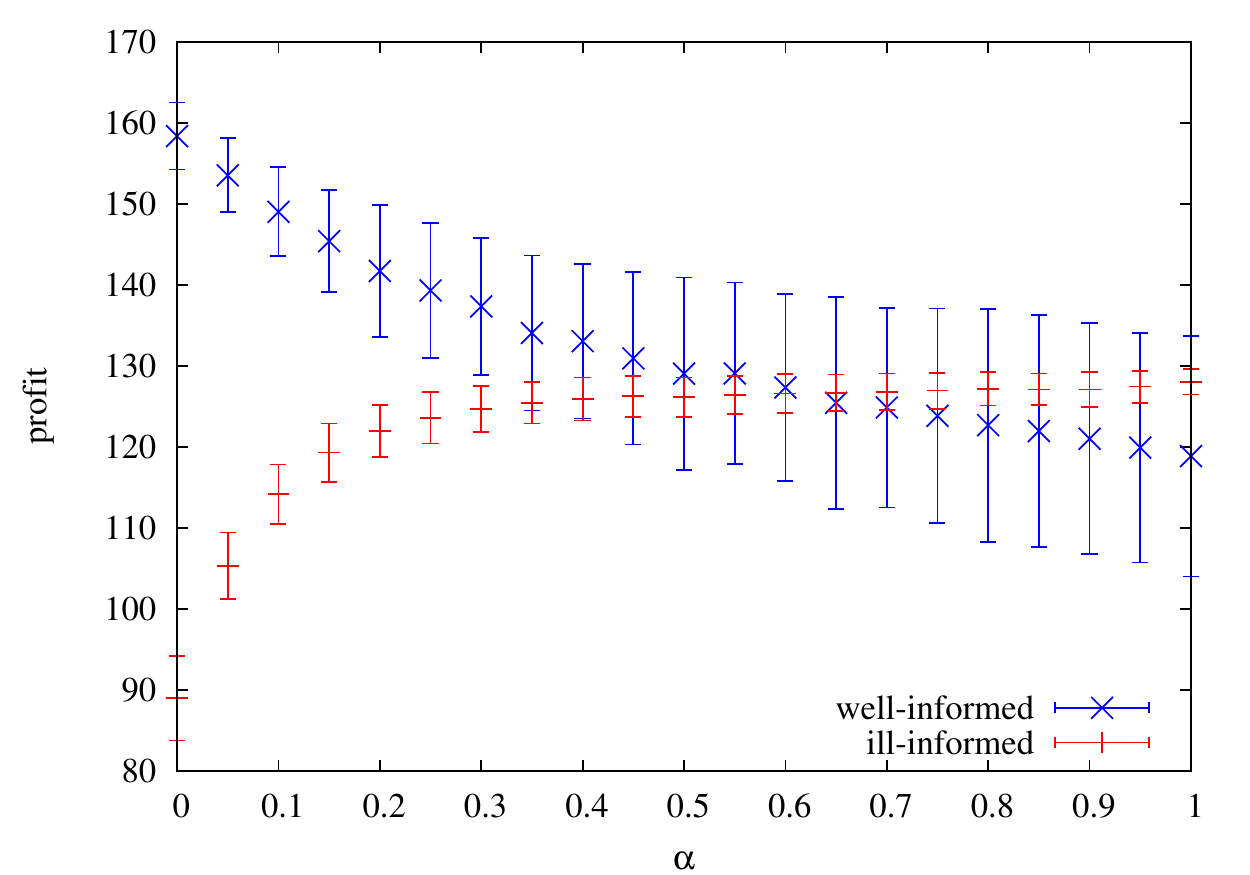}

 \end{itemize}
 
\pagebreak 
 \item Information rate $\Delta N_\text{arg}=3$:

 \begin{itemize}
	\item Fraction of \ills\ $f_{\ii}=0.1$:

	\leftFig{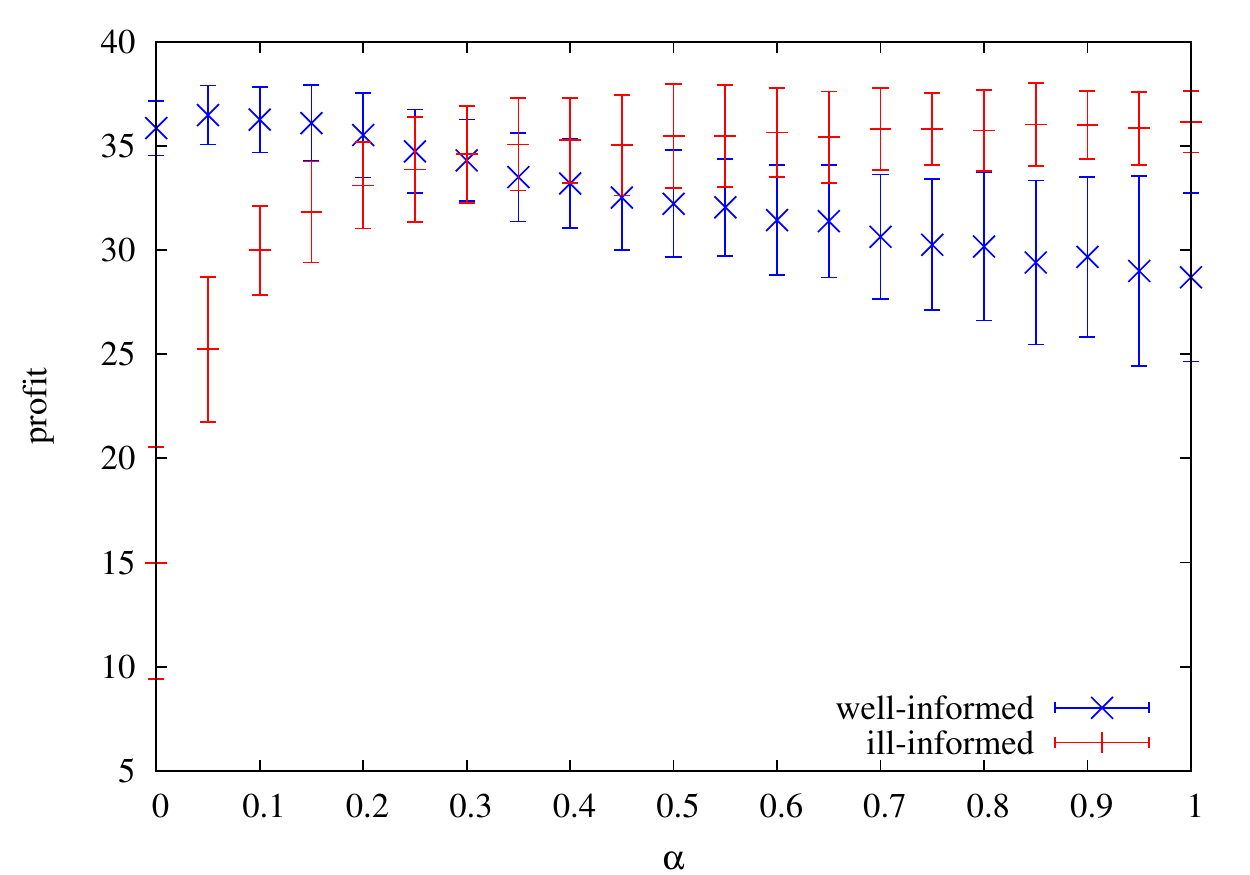}
	\rightFig{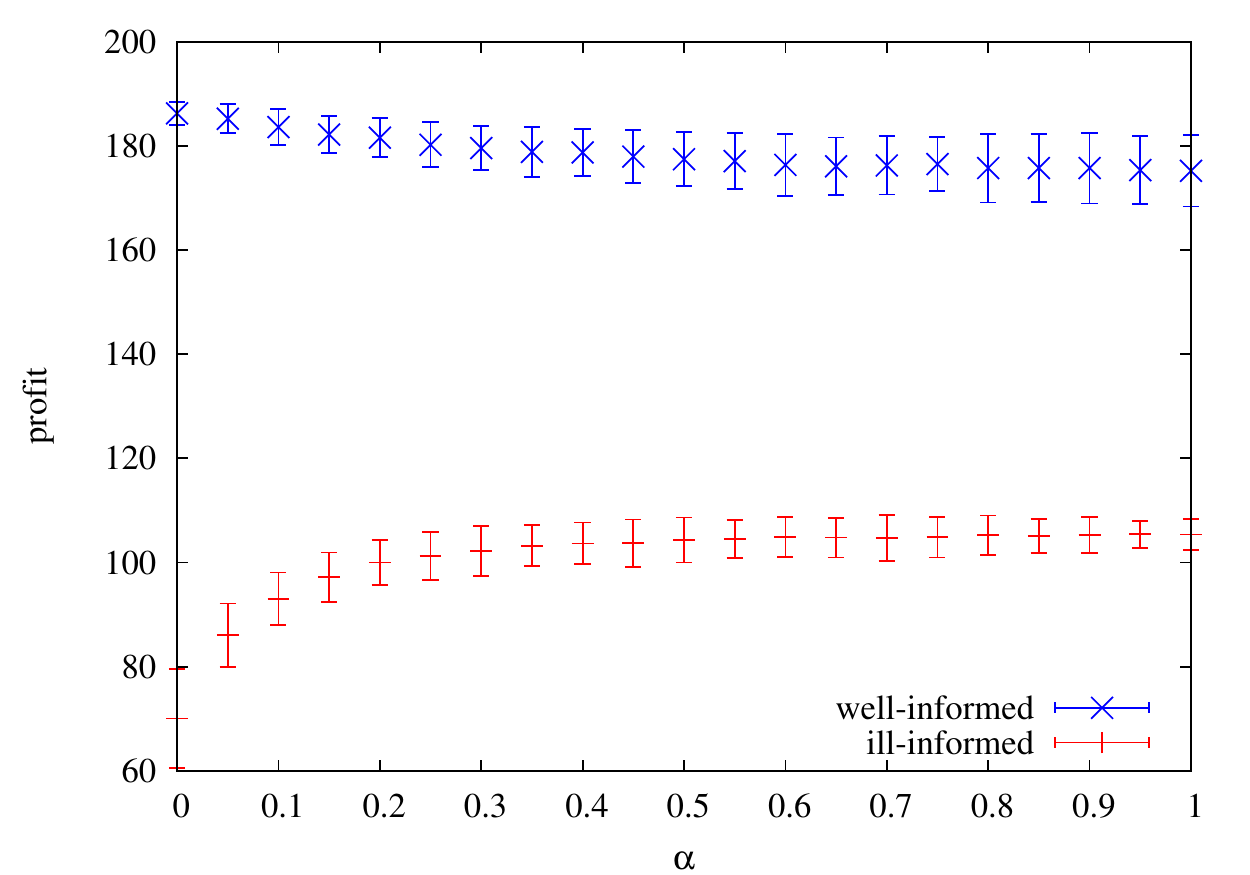}

	\item Fraction of \ills\ $f_{\ii}=0.5$
	
	\leftFig{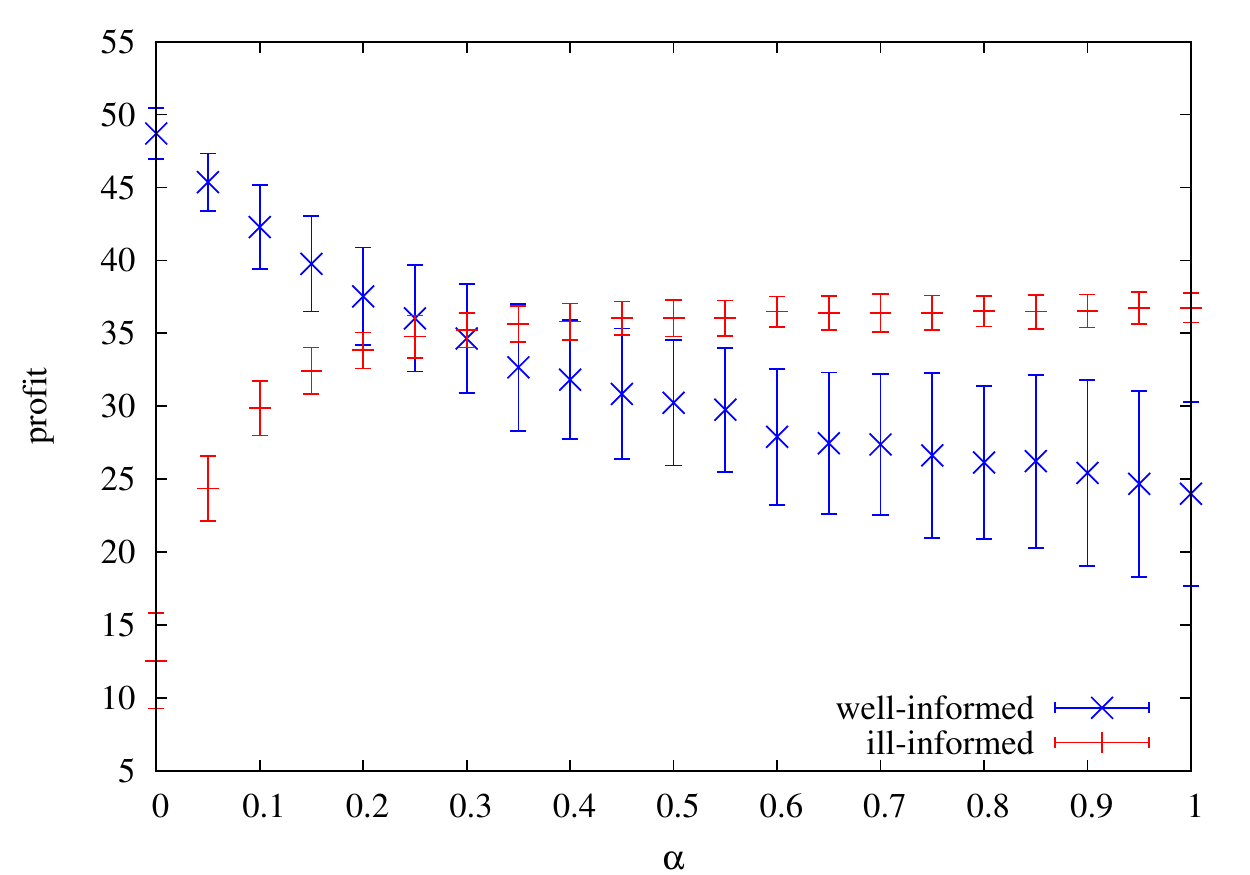}
	\rightFig{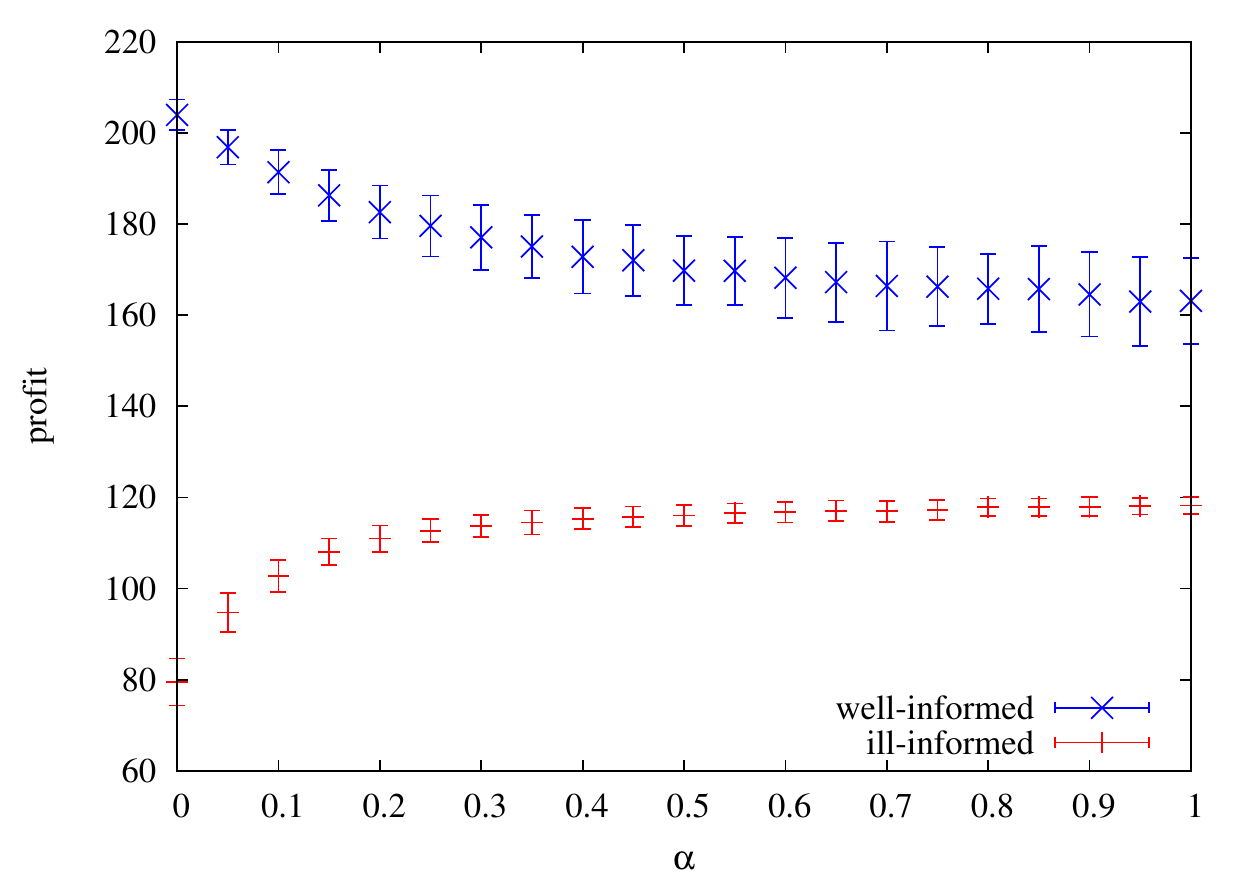}

	\item Fraction of \ills\ $f_{\ii}=0.9$

	\leftFig{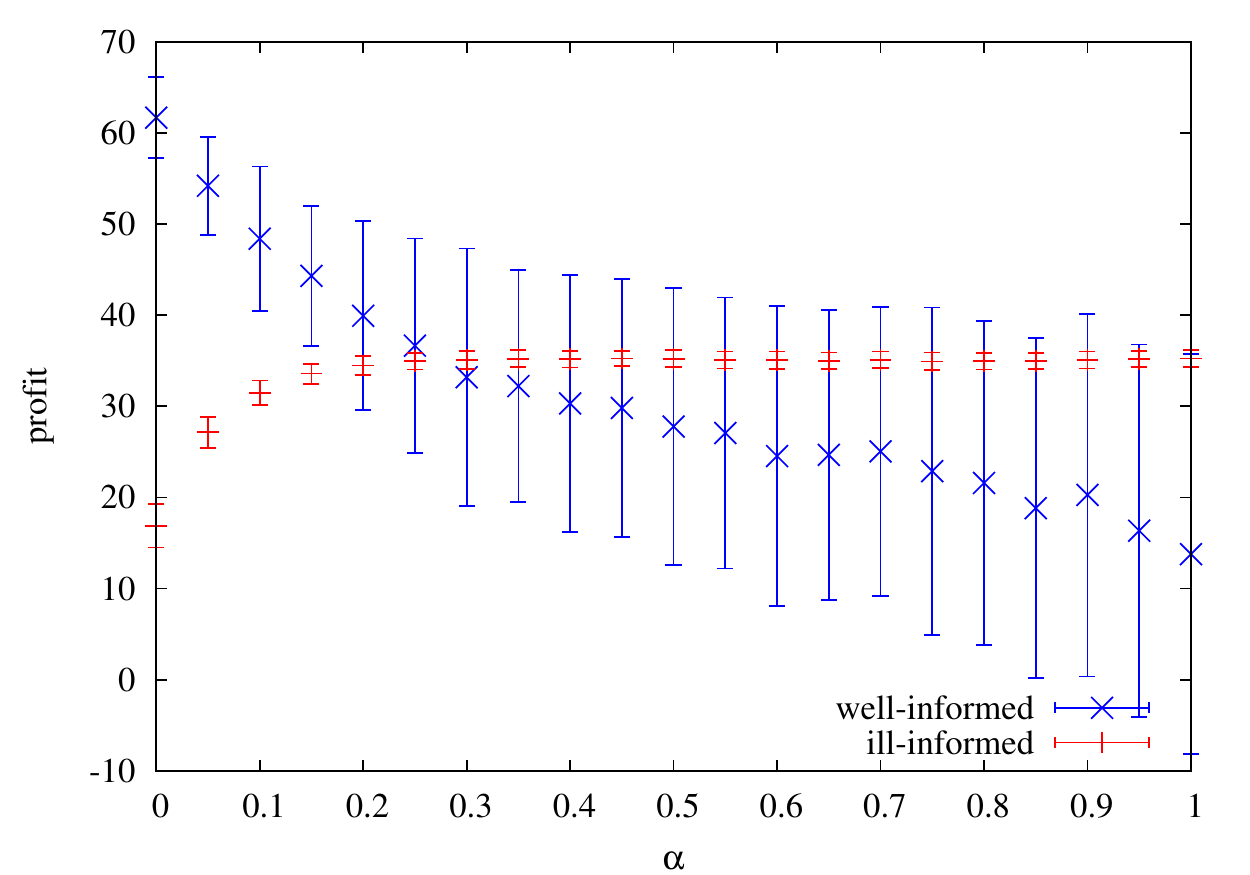}
	\rightFig{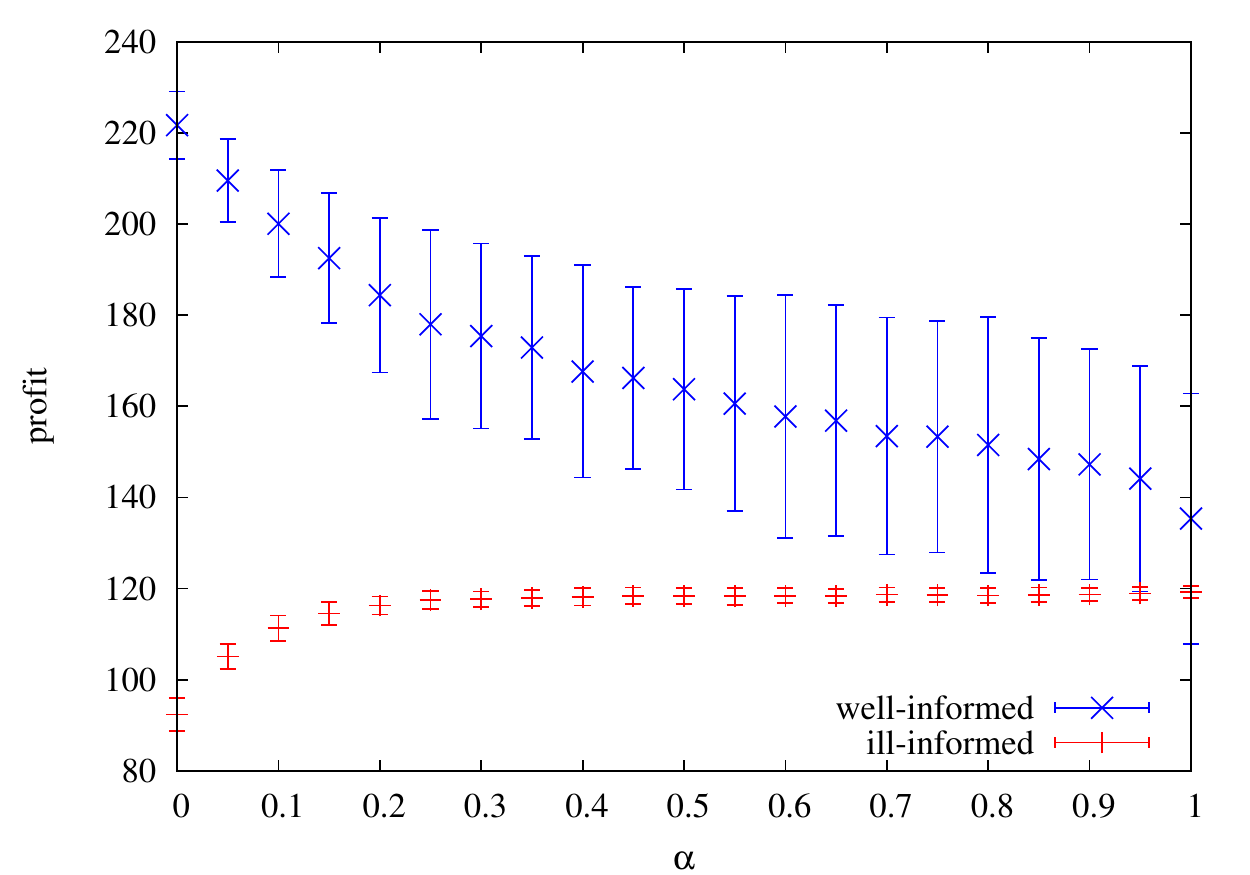}

 \end{itemize}

\pagebreak
 \item Information rate $\Delta N_\text{arg}=4$:

 \begin{itemize}
	\item Fraction of \ills\ $f_{\ii}=0.1$:

	\leftFig{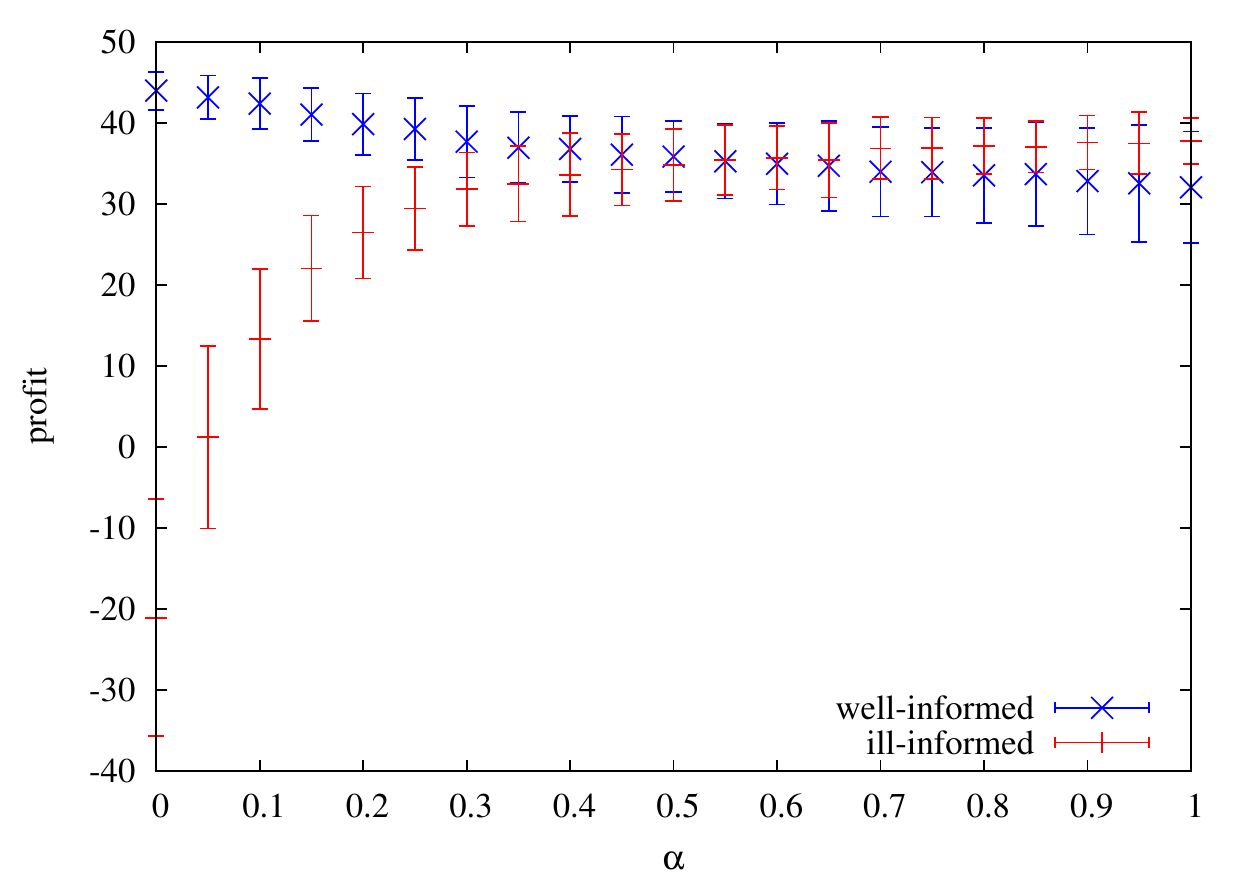}
	\rightFig{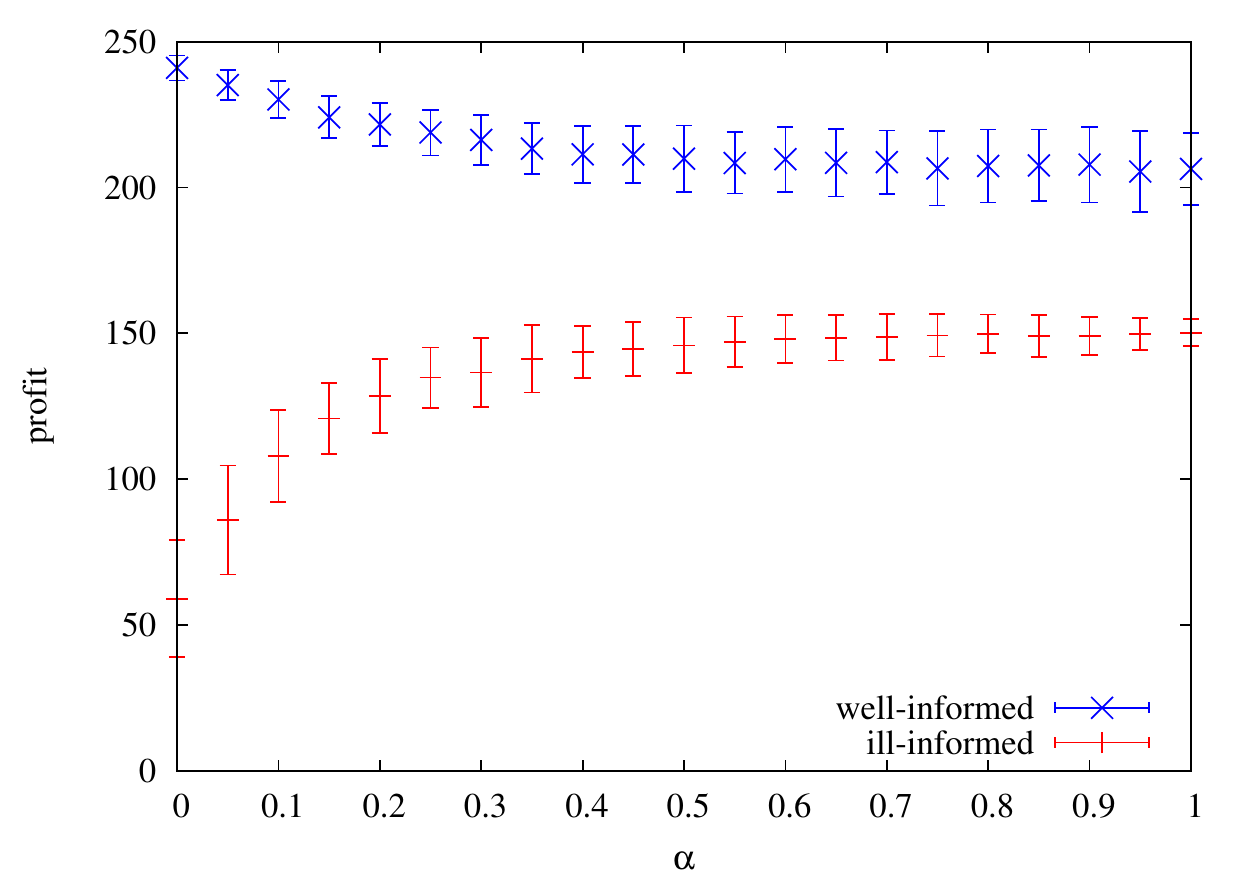}

	\item Fraction of \ills\ $f_{\ii}=0.5$
	
	\leftFig{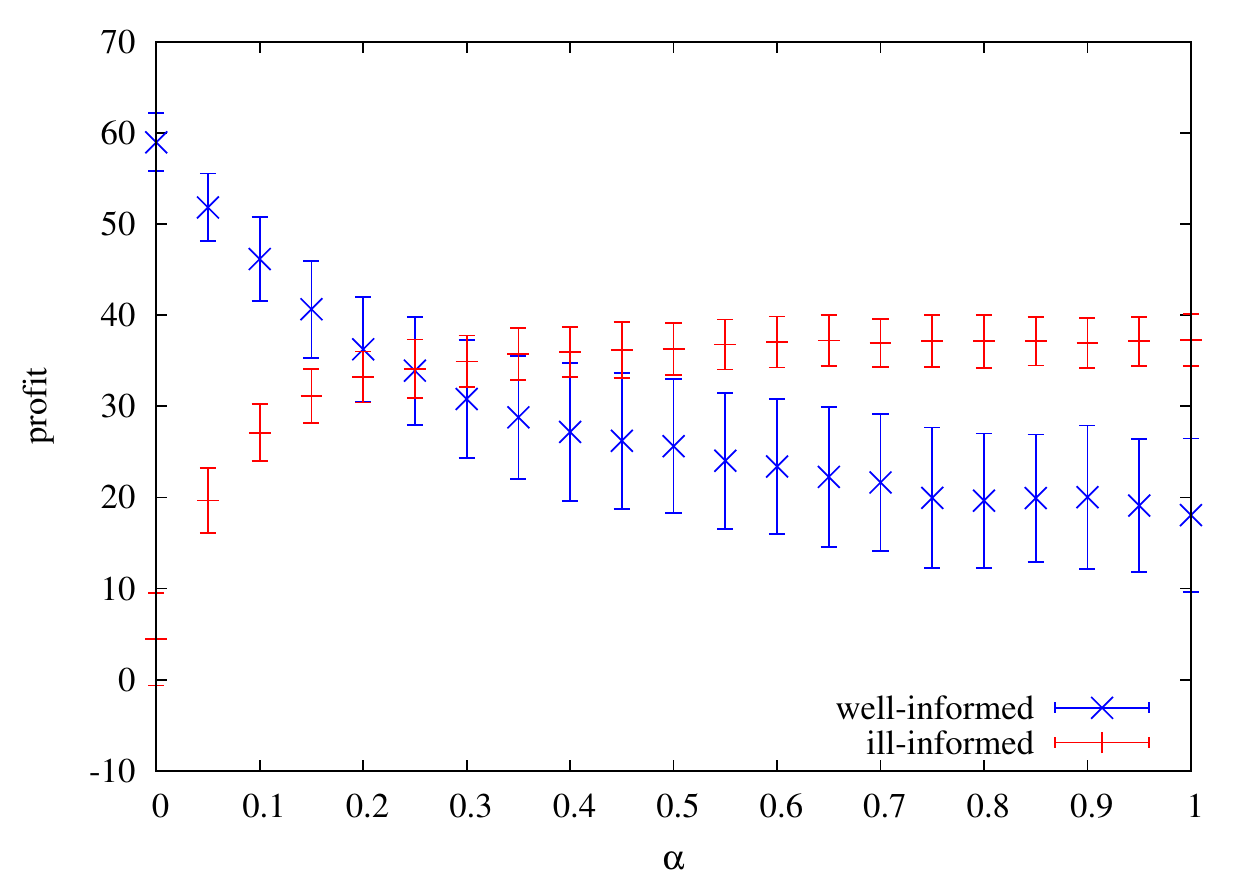}
	\rightFig{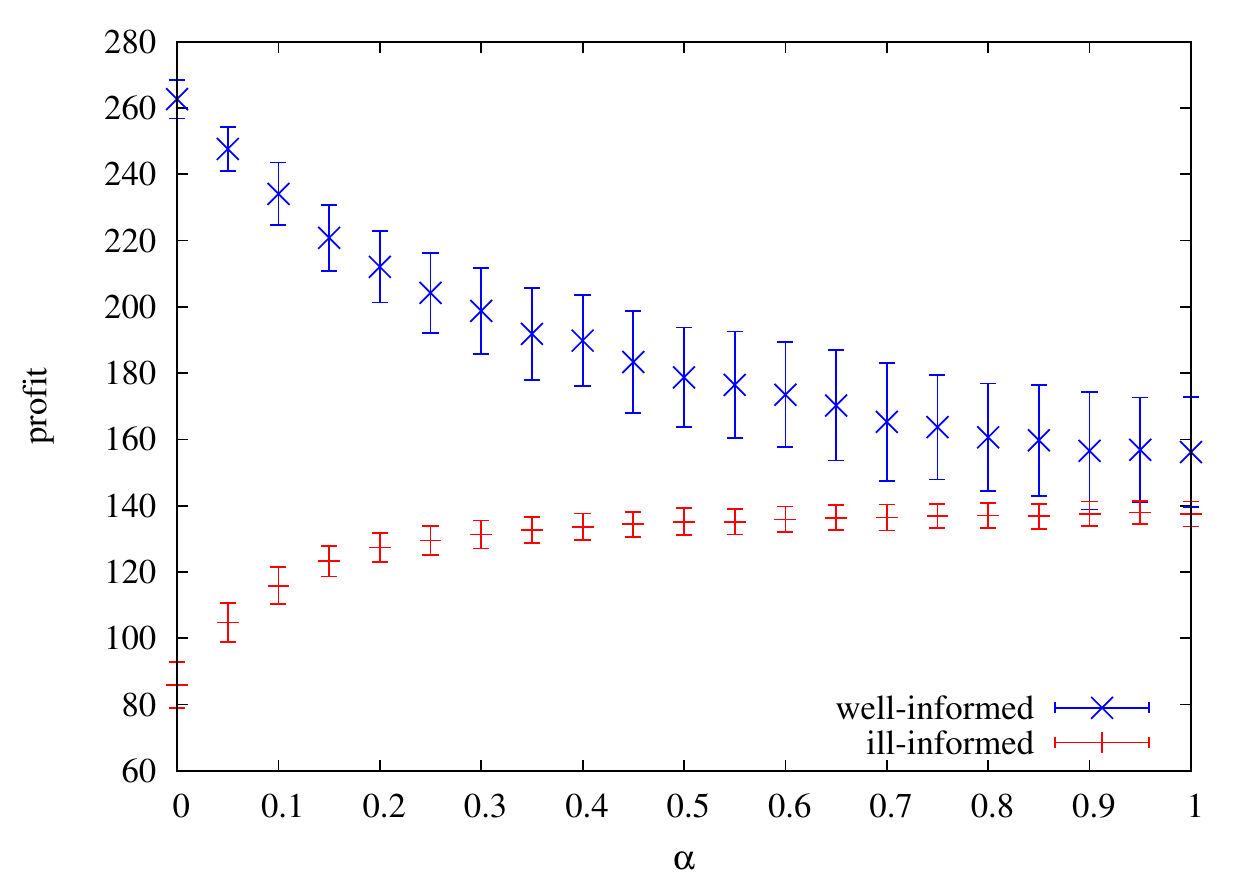}

	\item Fraction of \ills\ $f_{\ii}=0.9$

	\leftFig{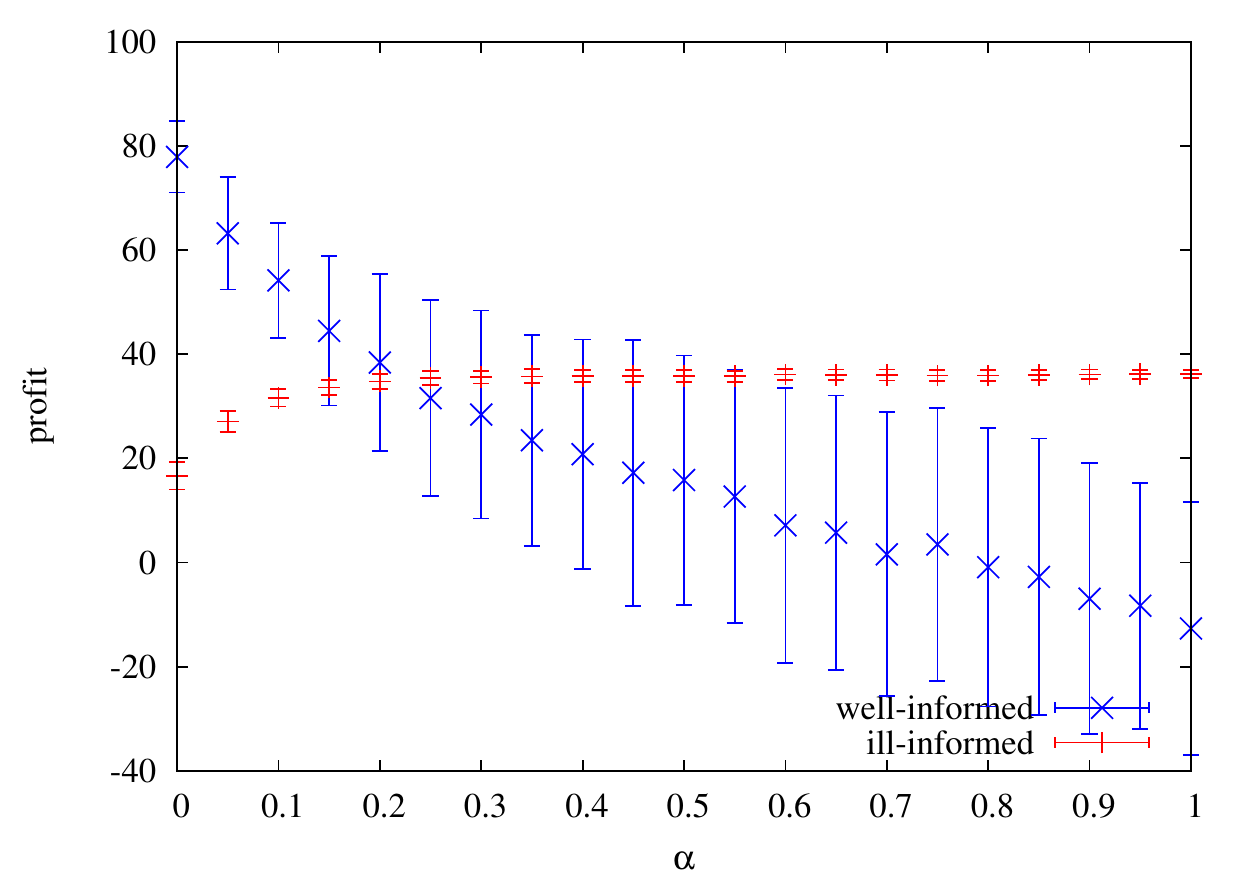}
	\rightFig{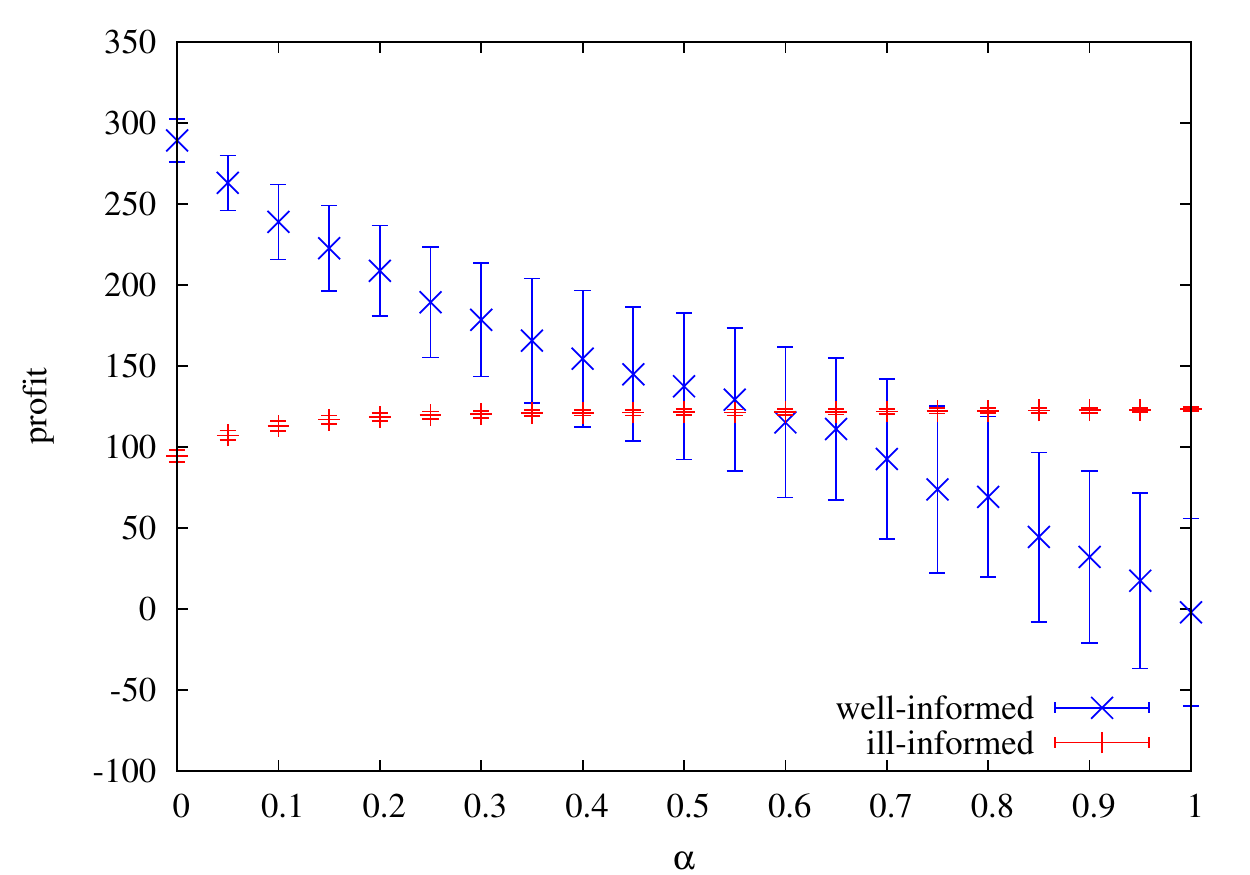}

 \end{itemize}

\end{itemize}

%\pagebreak
%\paragraph{Varying Number Of Arguments At Beggining}
%
%We noticed that the initial number of arguments given to the consultants has an influence on their payoffs. 
%This is because when more arguments are given to \ills, they can build their reputation in the beginning.
%Although we think that for this reason, a number of is 

\pagebreak
\subsubsection{Using R2}

\begin{itemize}
 
 \item Information rate $\Delta N_\text{arg}=2$:

 \begin{itemize}
	\item Fraction of \ills\ $f_{\ii}=0.1$:

	\leftFig{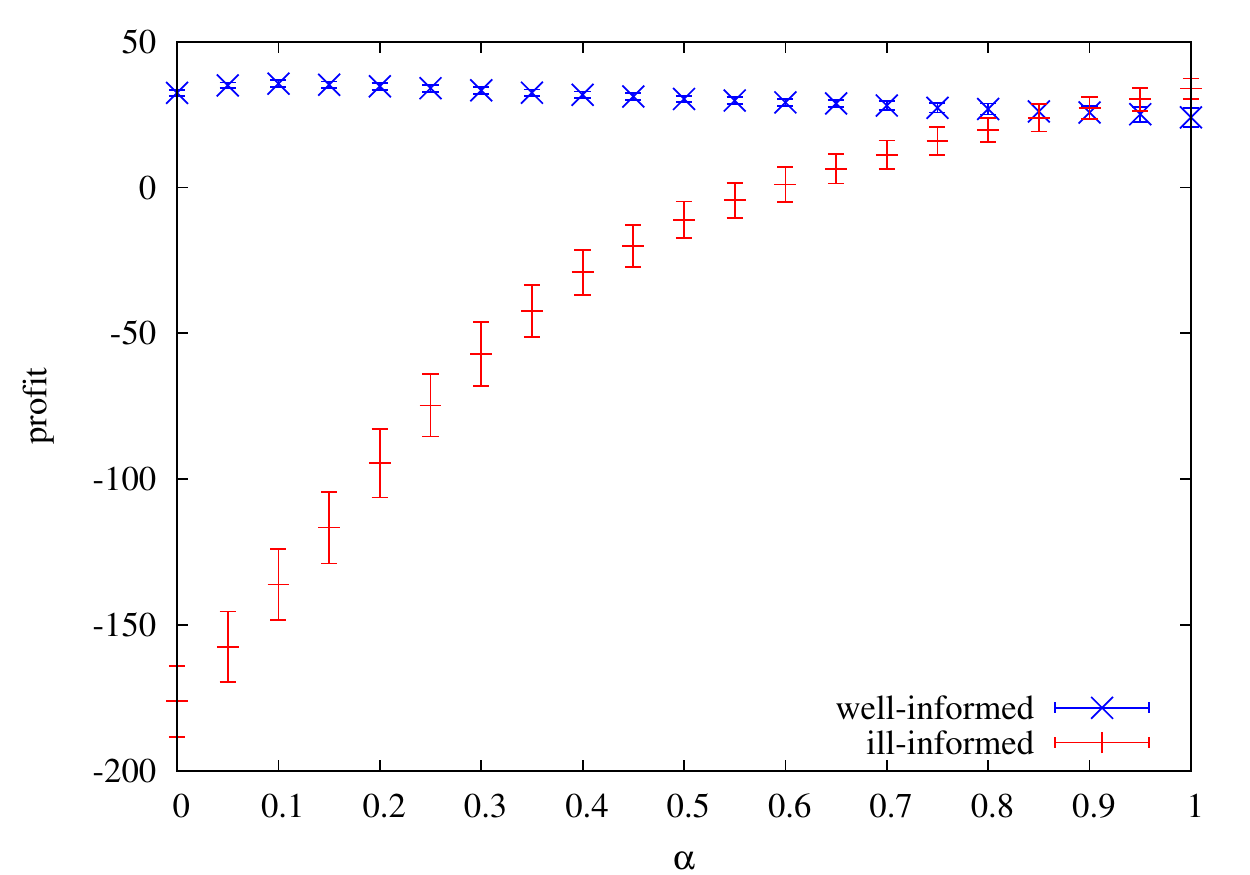}
	\rightFig{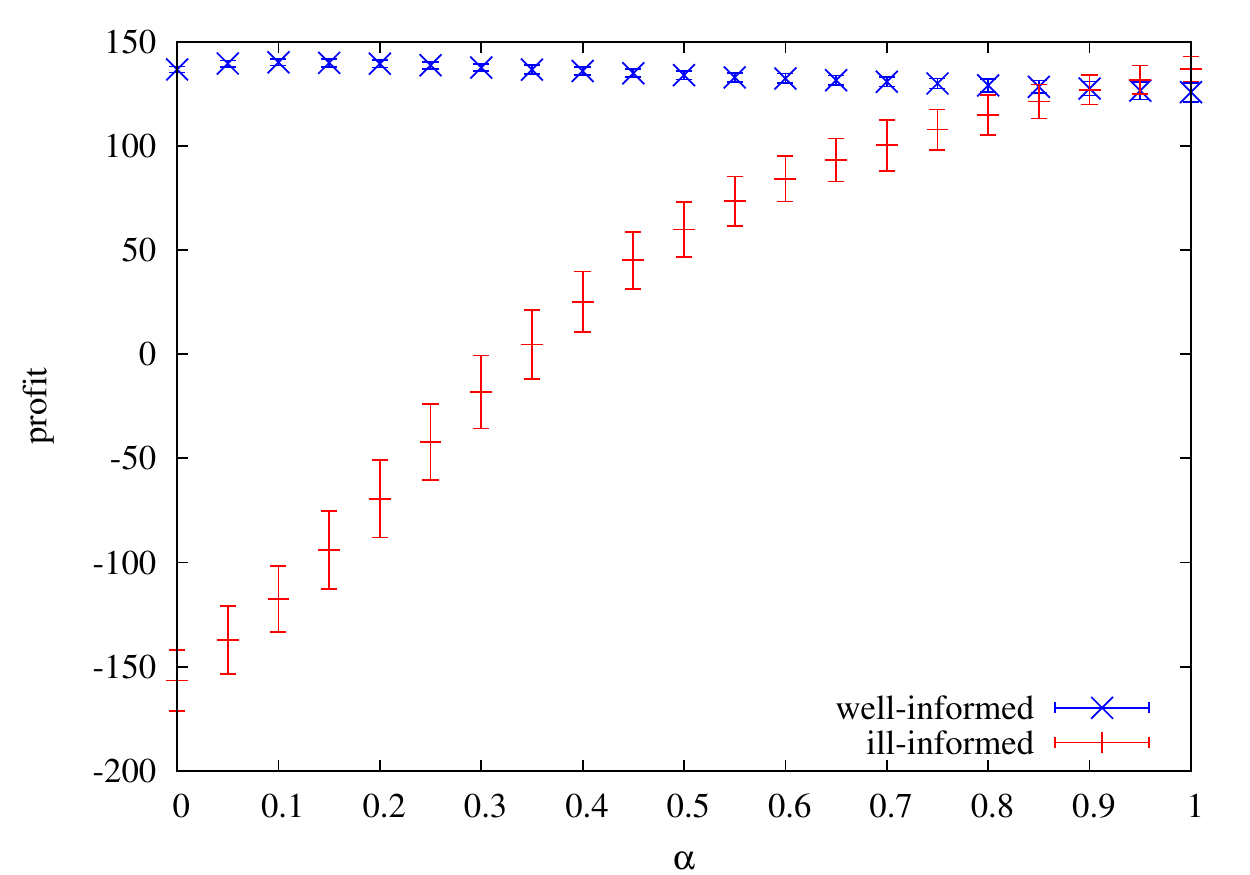}

	\item Fraction of \ills\ $f_{\ii}=0.5$
	
	\leftFig{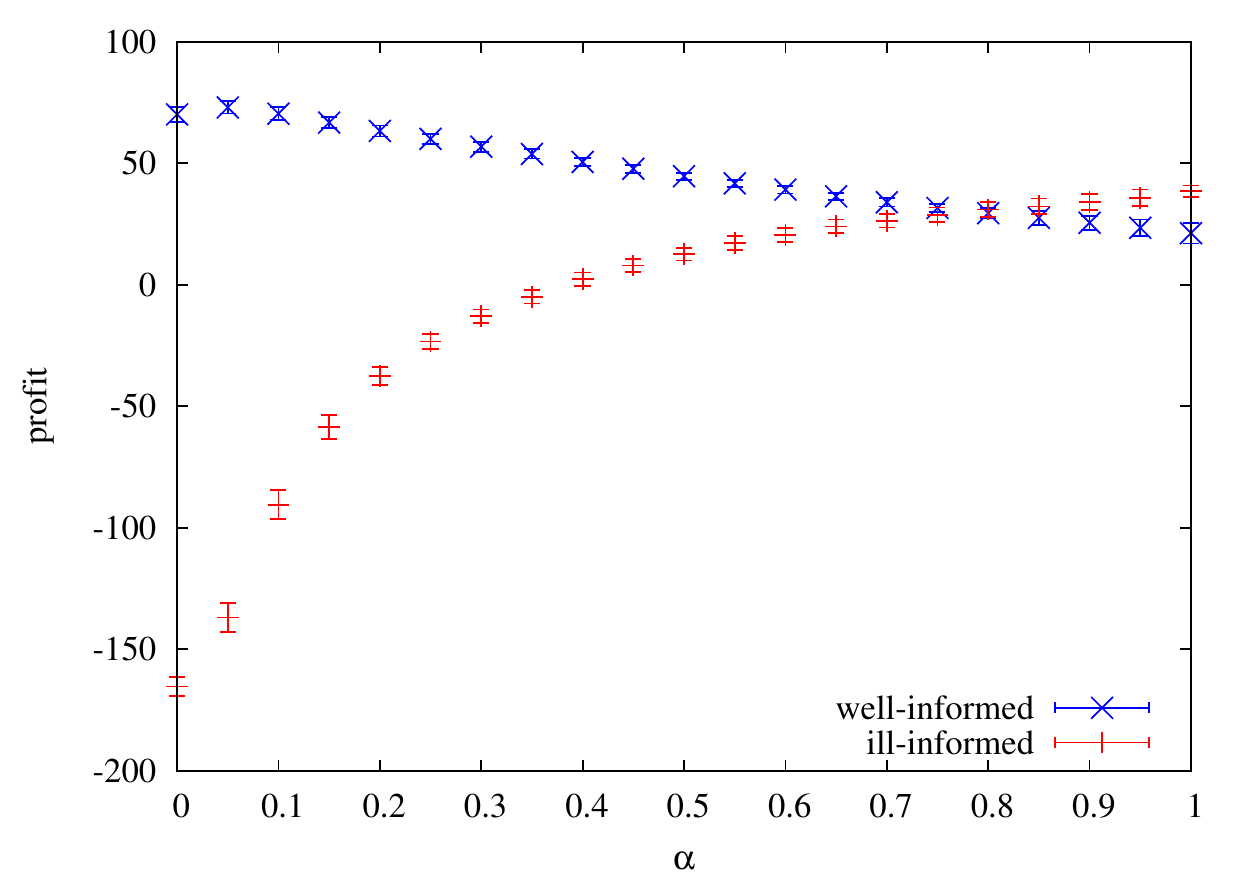}
	\rightFig{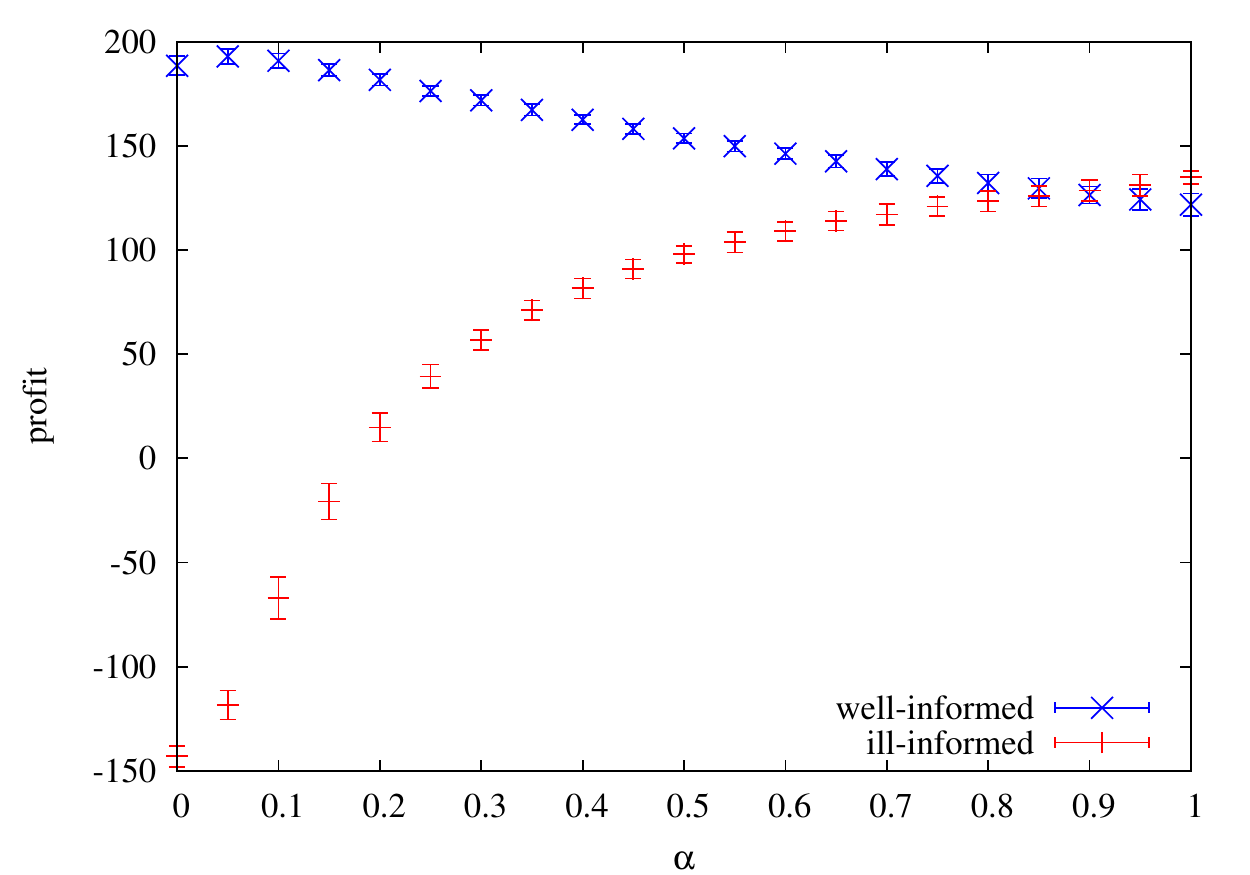}

	\item Fraction of \ills\ $f_{\ii}=0.9$

	\leftFig{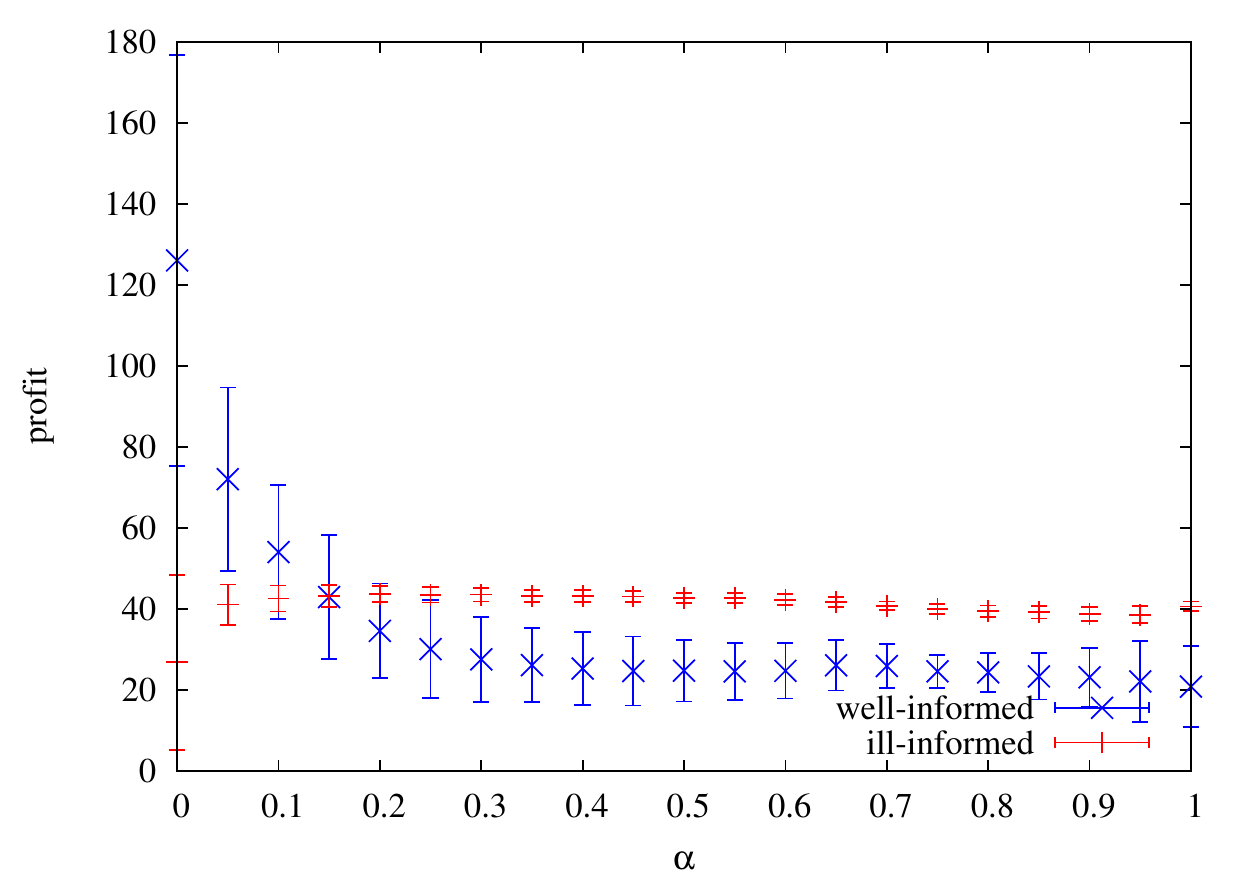}
	\rightFig{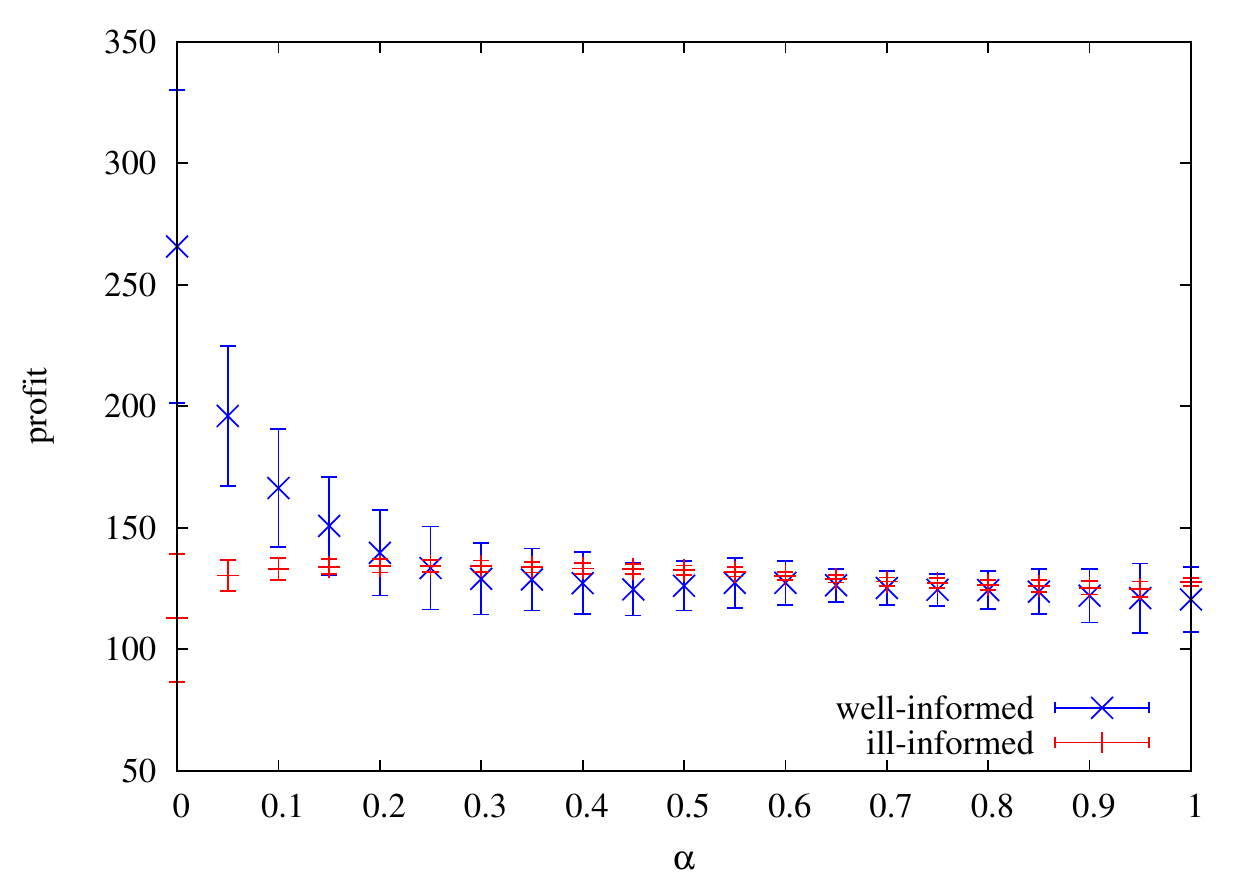}

 \end{itemize}
 
\pagebreak 
 \item Information rate $\Delta N_\text{arg}=3$:

 \begin{itemize}
	\item Fraction of \ills\ $f_{\ii}=0.1$:

	\leftFig{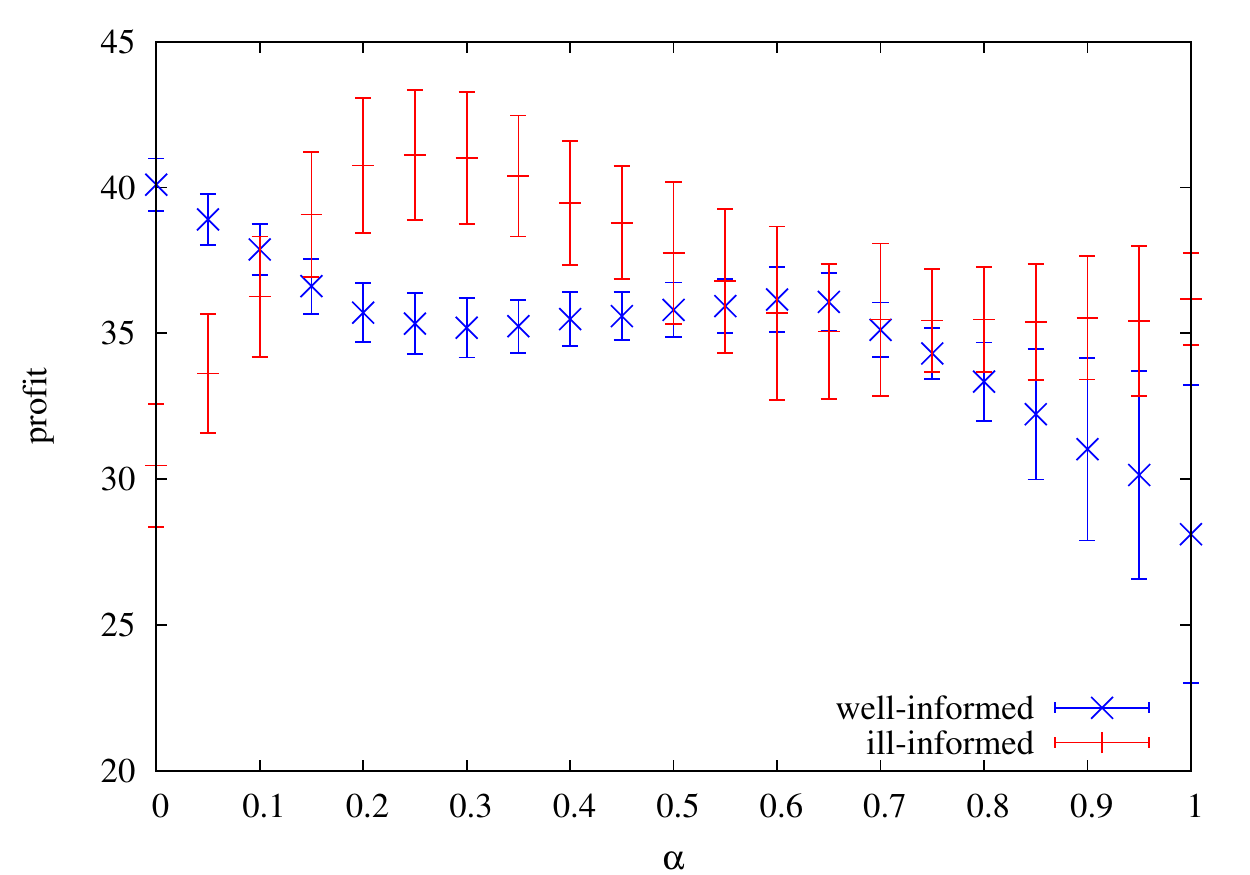}
	\rightFig{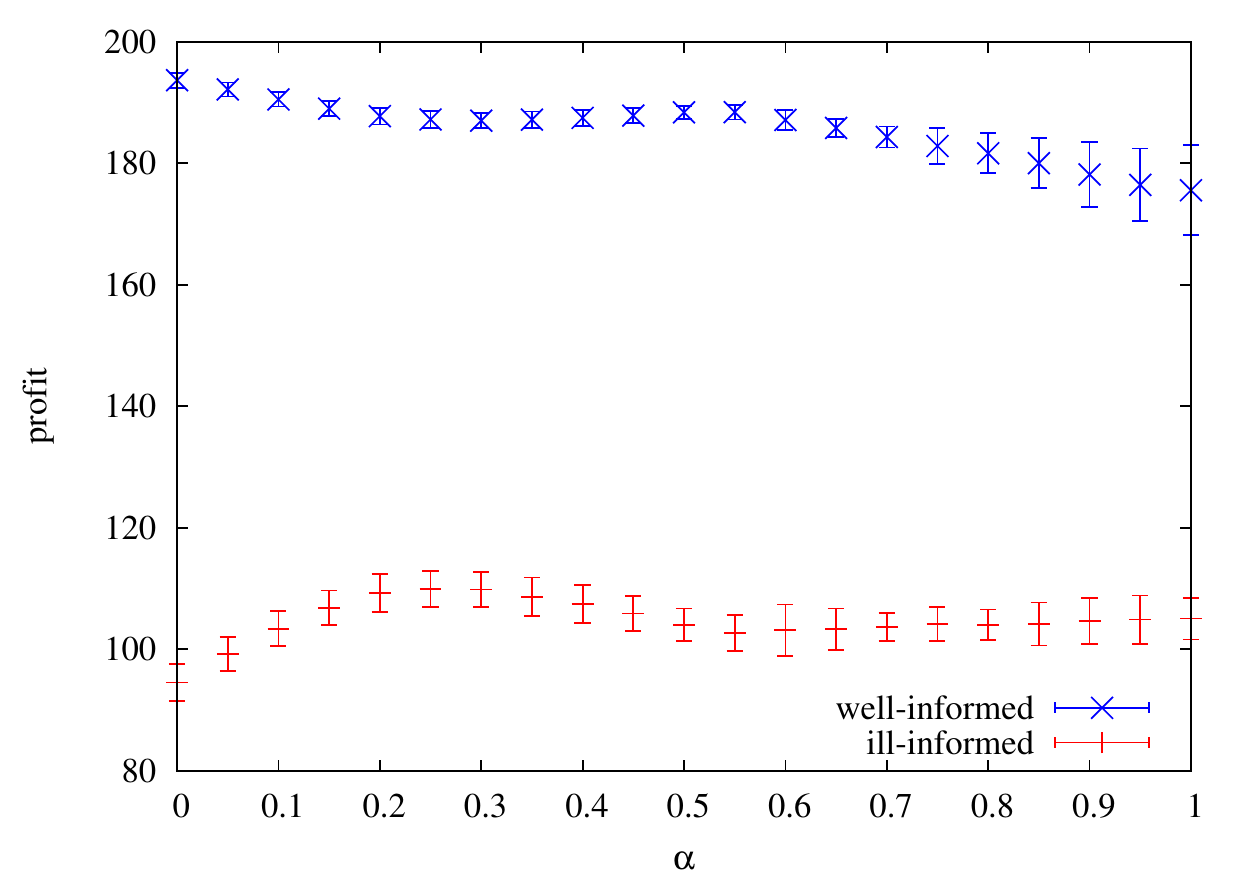}

	\item Fraction of \ills\ $f_{\ii}=0.5$
	
	\leftFig{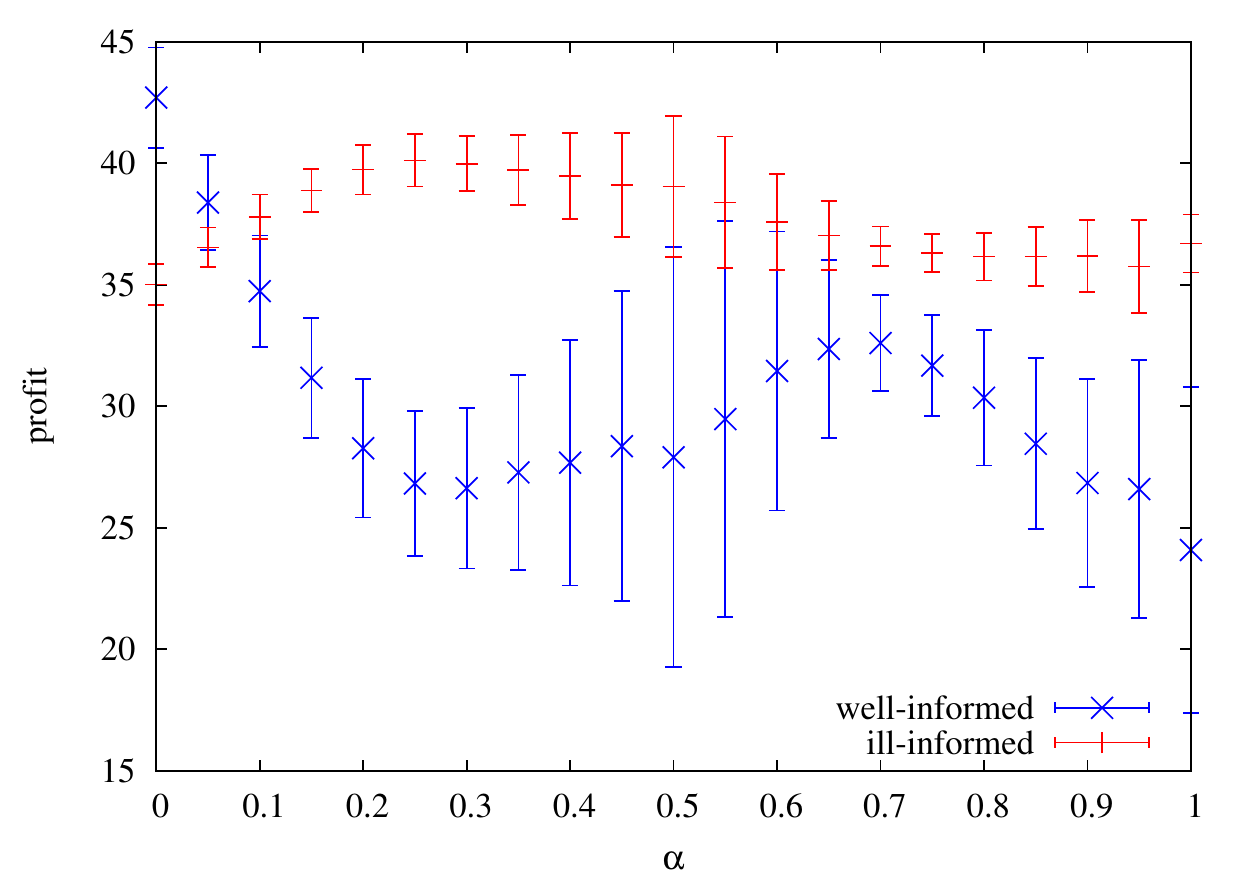}
	\rightFig{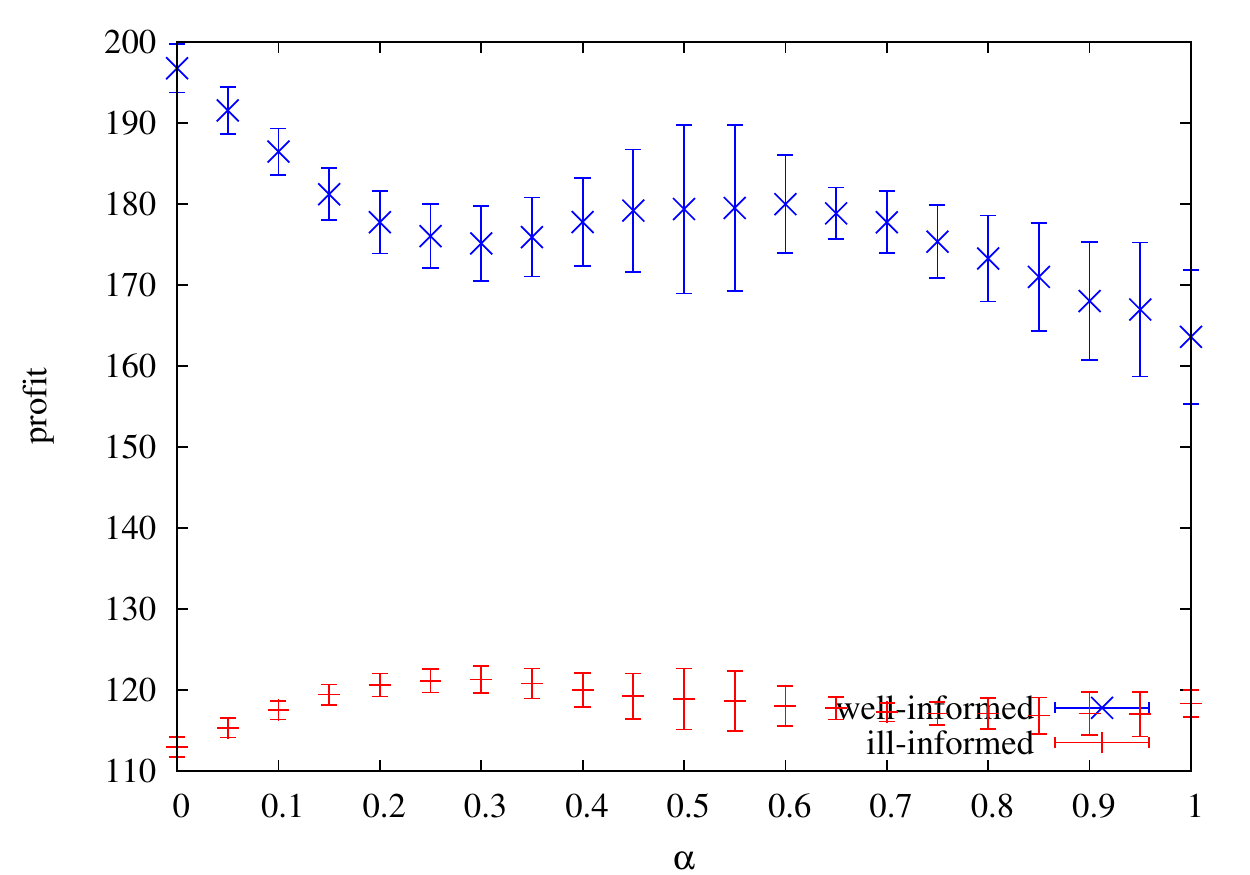}

	\item Fraction of \ills\ $f_{\ii}=0.9$

	\leftFig{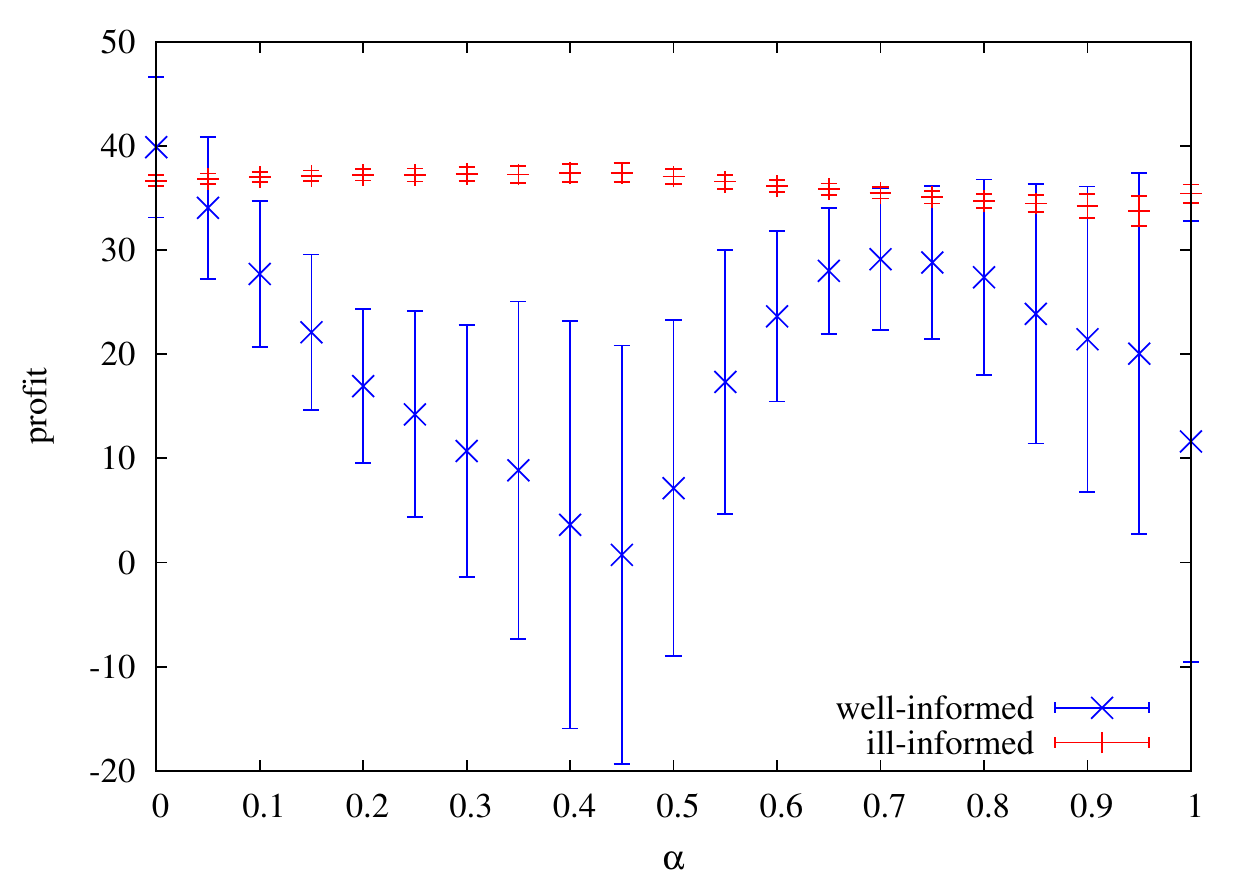}
	\rightFig{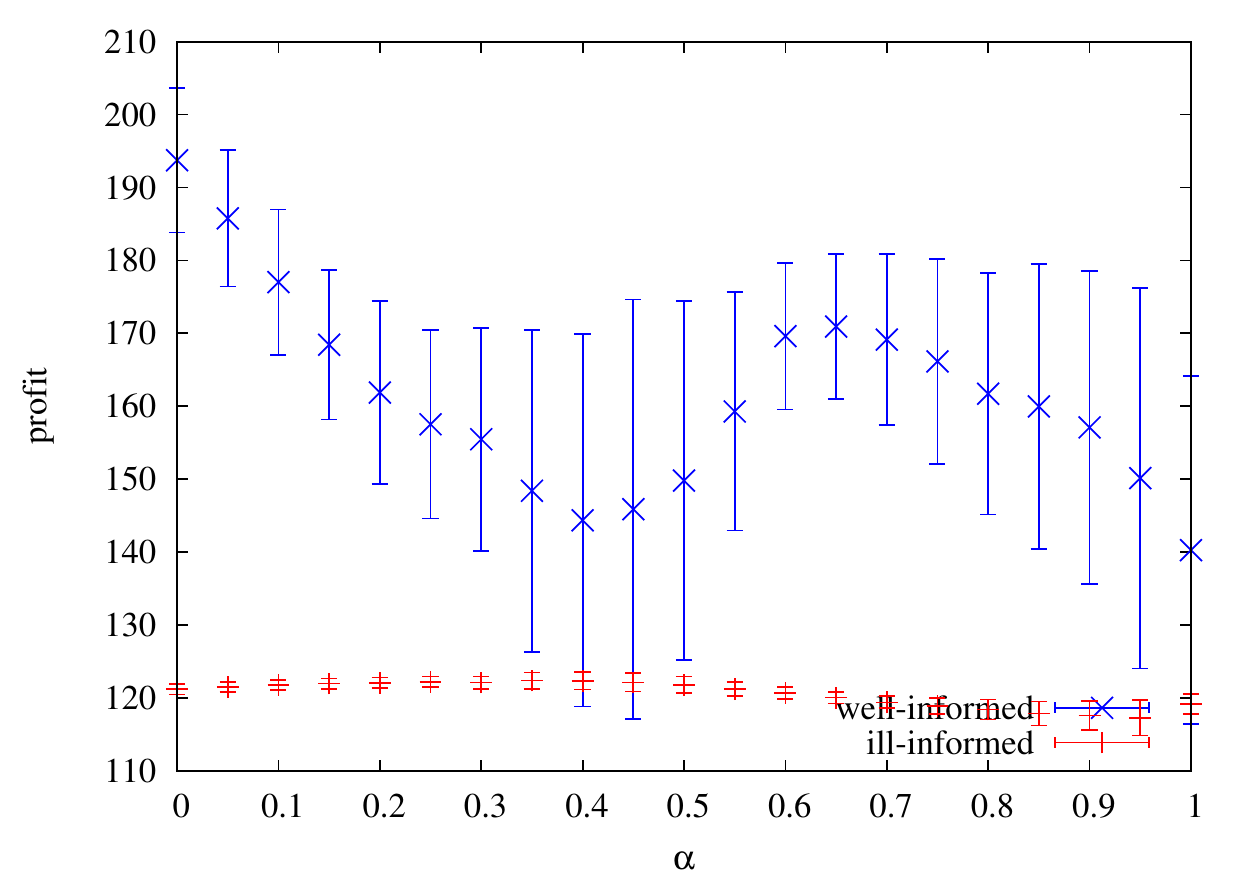}

 \end{itemize}

\pagebreak
 \item Information rate $\Delta N_\text{arg}=4$:

 \begin{itemize}
	\item Fraction of \ills\ $f_{\ii}=0.1$:

	\leftFig{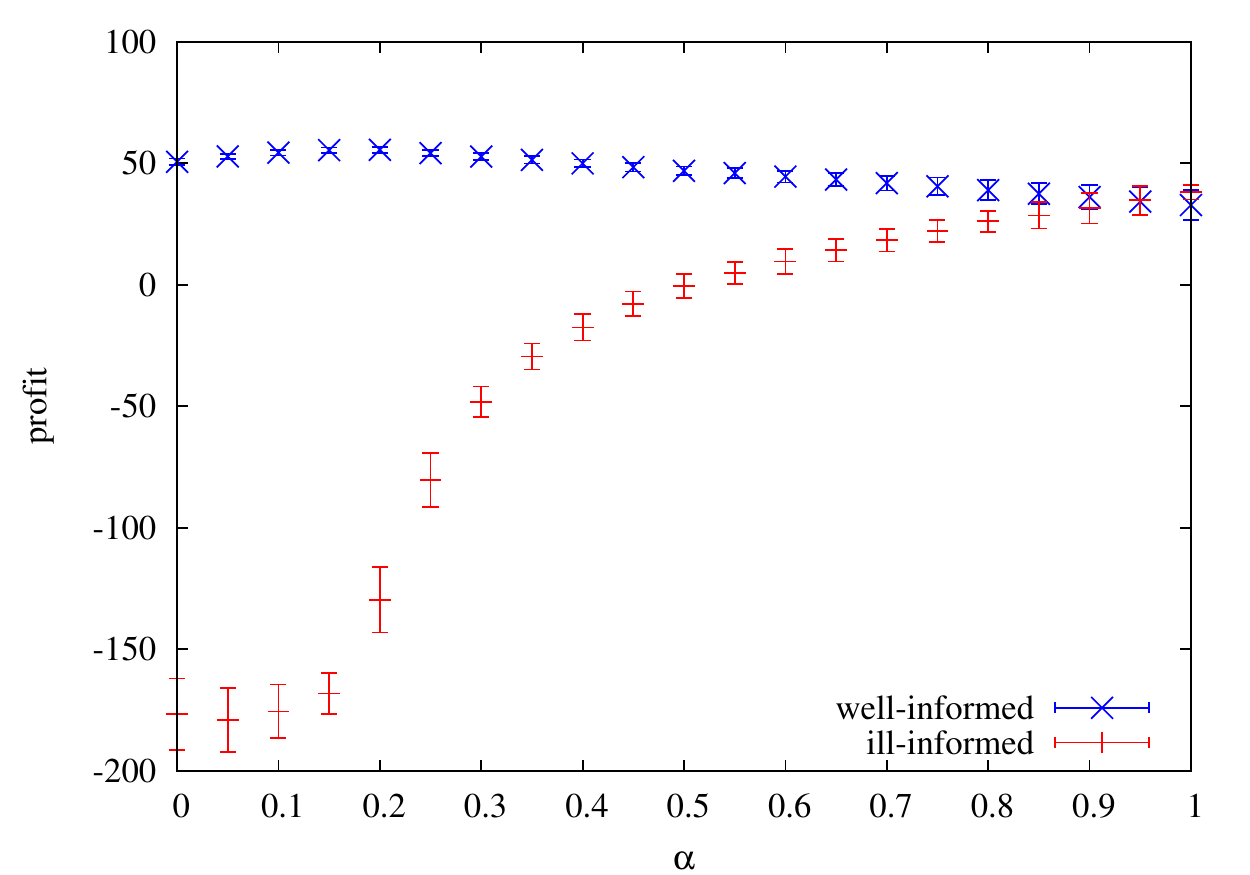}
	\rightFig{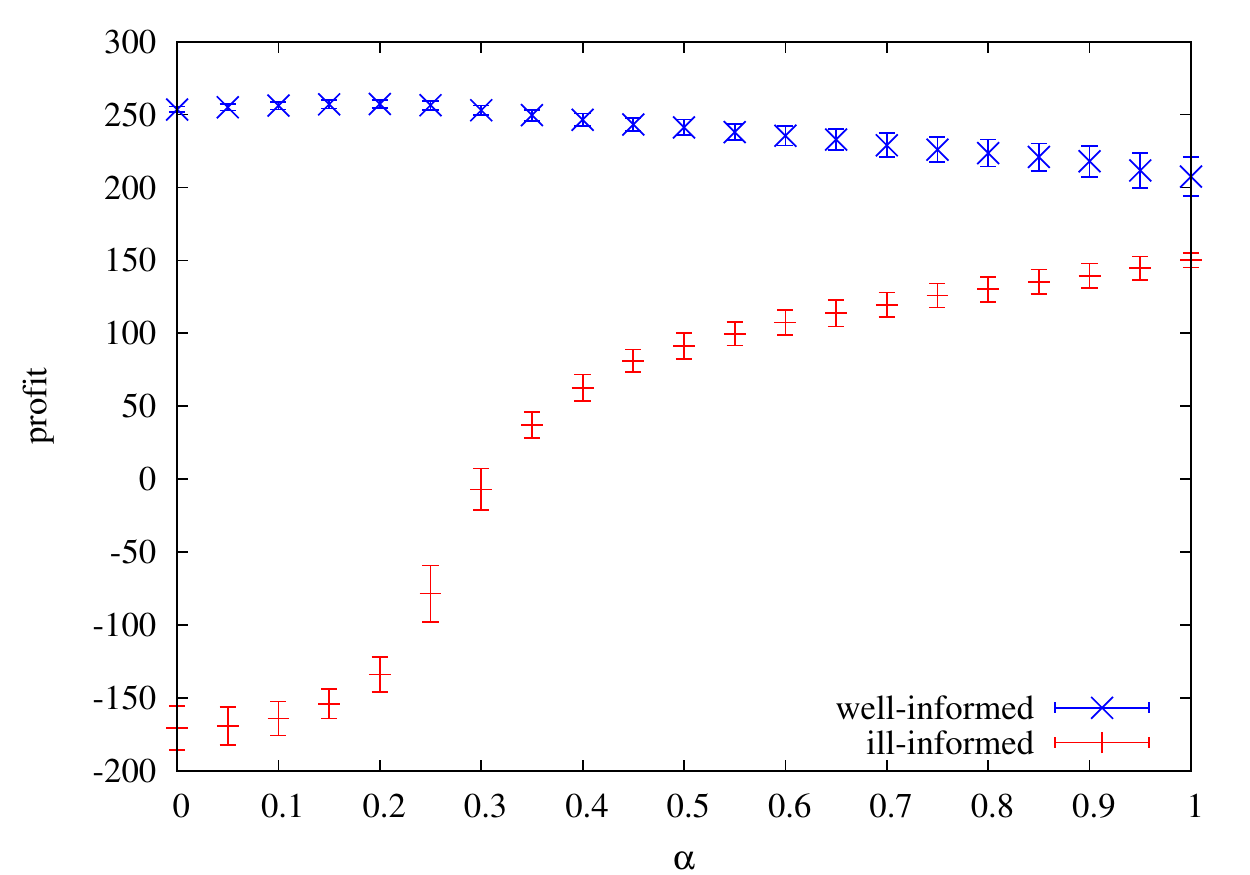}

	\item Fraction of \ills\ $f_{\ii}=0.5$
	
	\leftFig{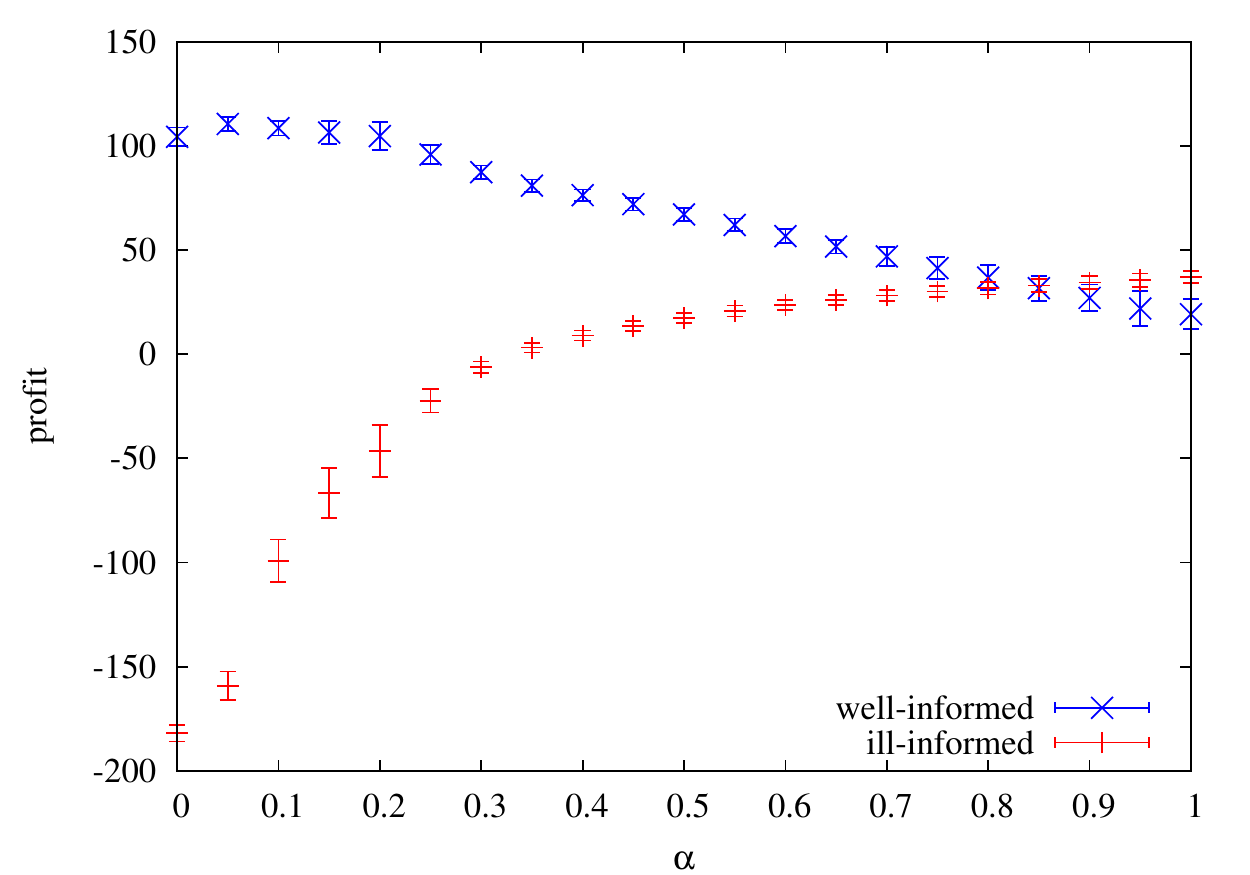}
	\rightFig{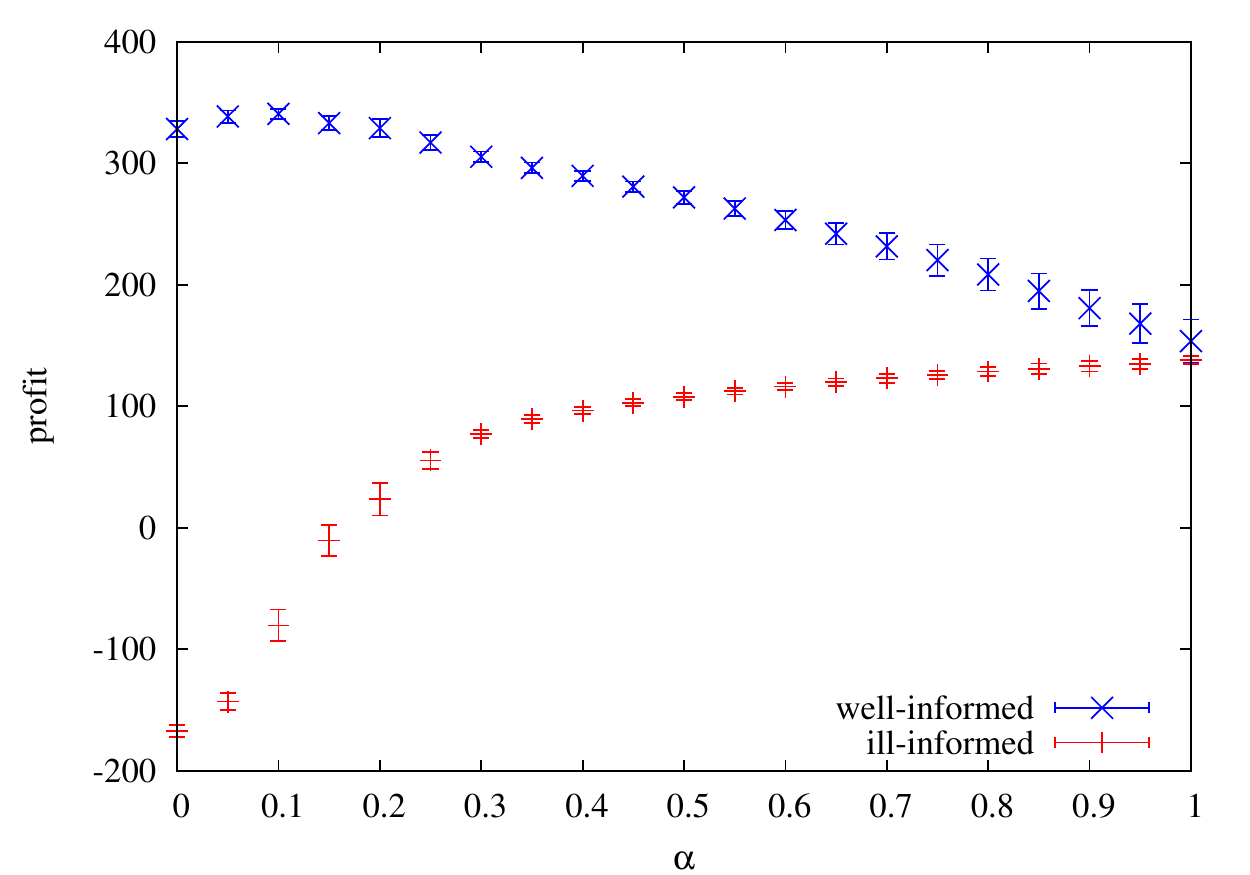}

	\item Fraction of \ills\ $f_{\ii}=0.9$

	\leftFig{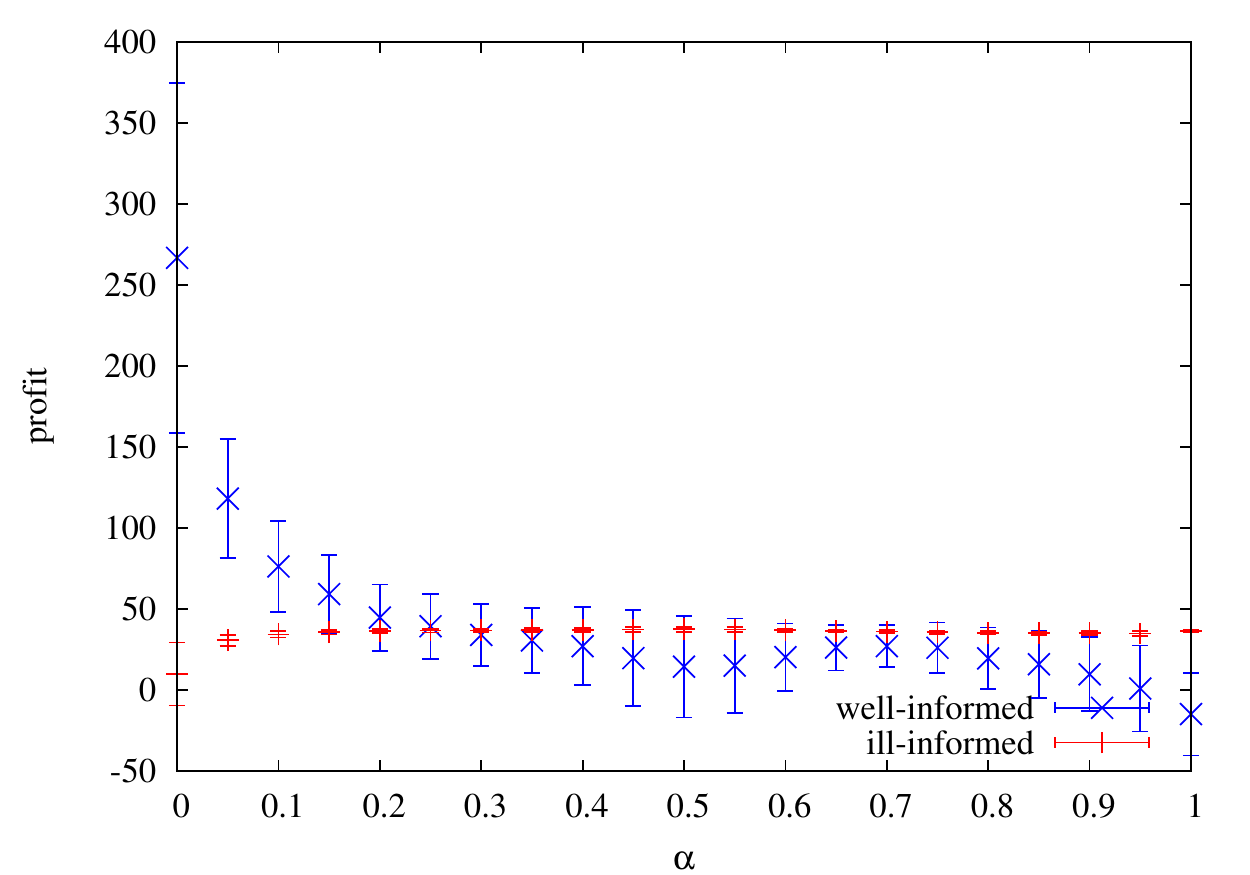}
	\rightFig{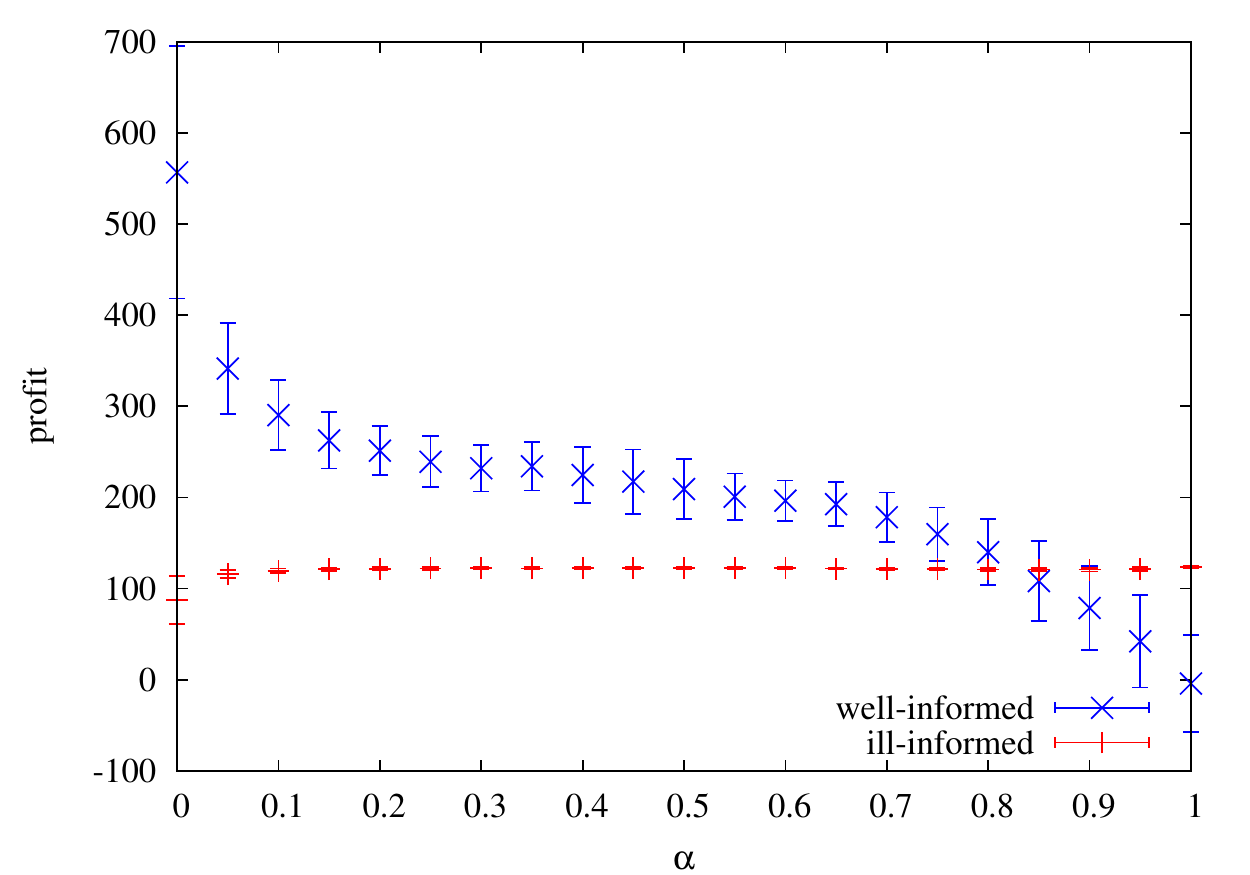}

 \end{itemize}

\pagebreak
 \item Information rate $\Delta N_\text{arg}=5$:

 \begin{itemize}
	\item Fraction of \ills\ $f_{\ii}=0.1$:

	\leftFig{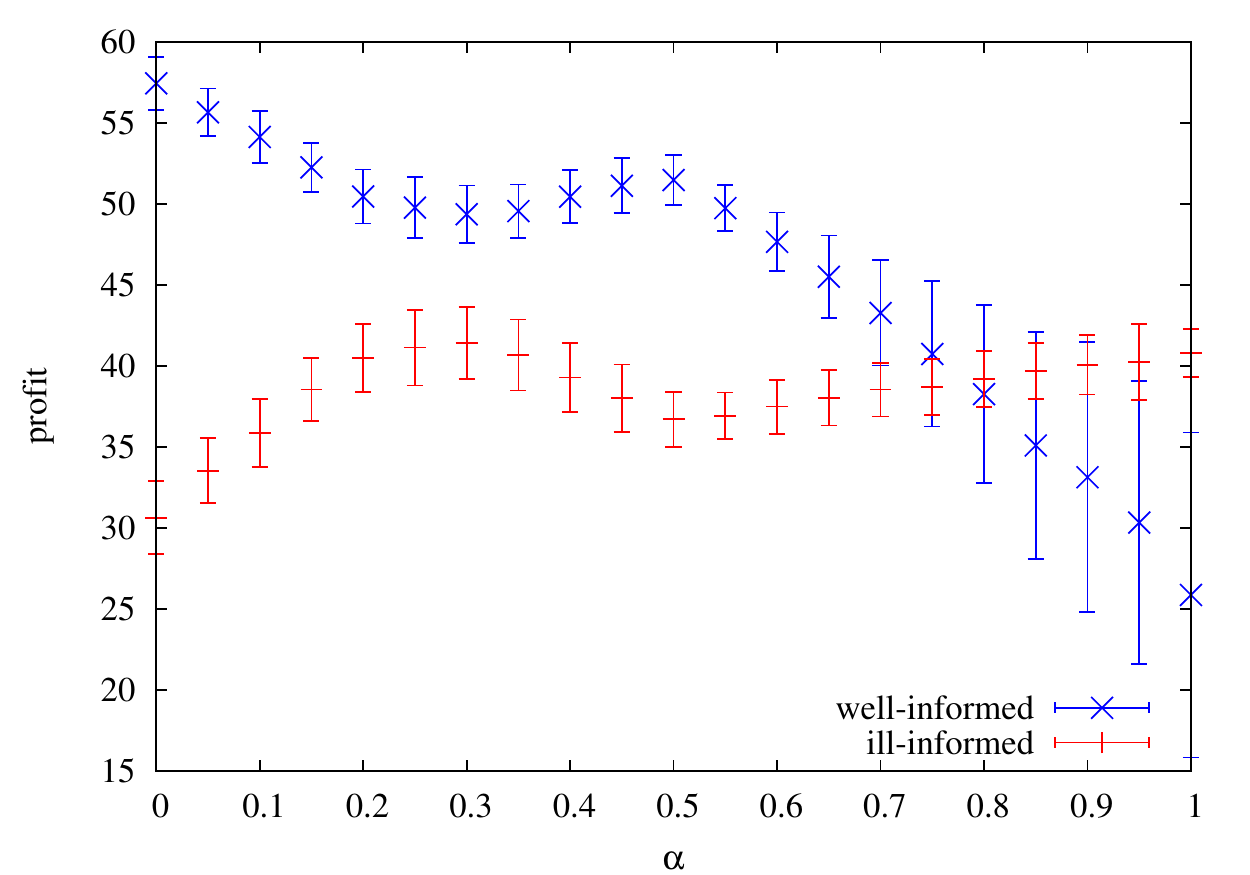}
	\rightFig{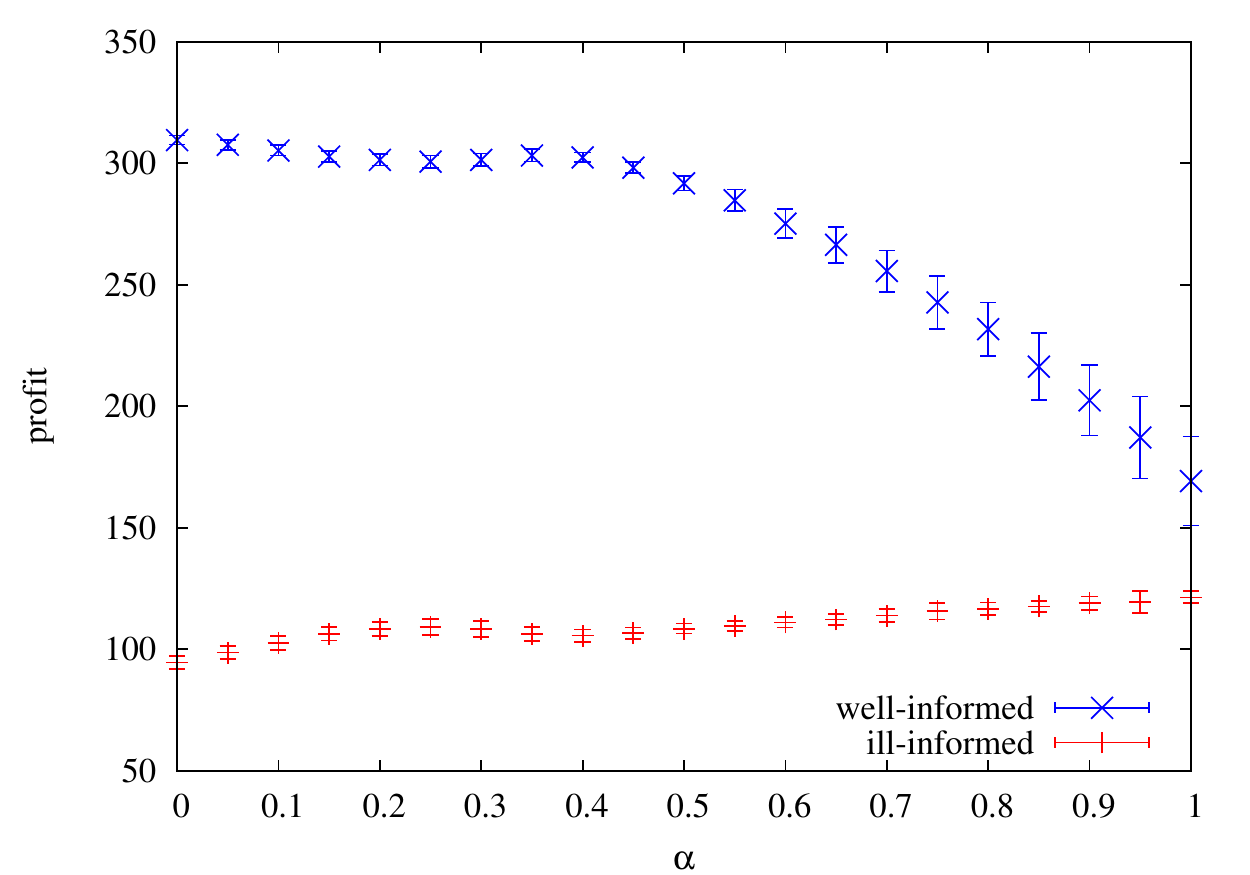}

	\item Fraction of \ills\ $f_{\ii}=0.5$
	
	\leftFig{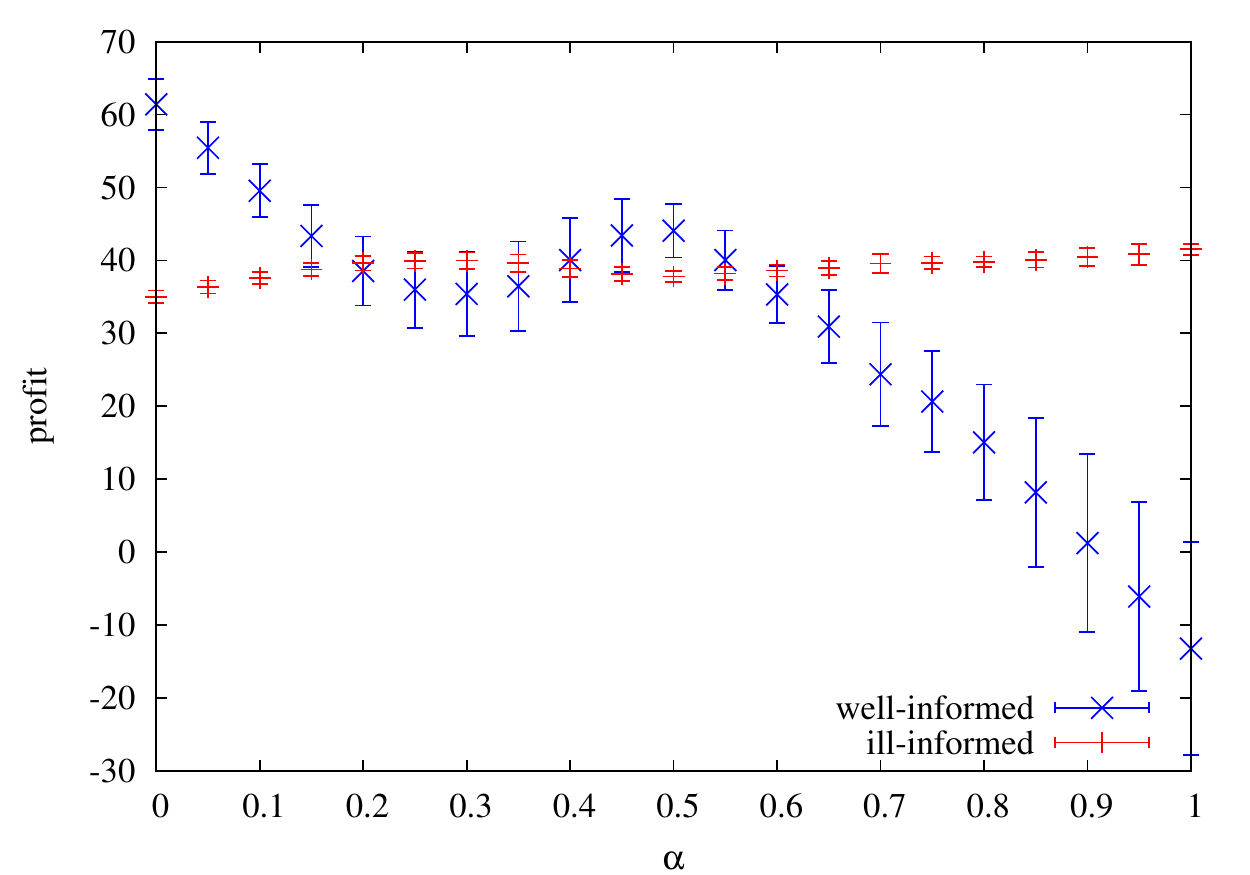}
	\rightFig{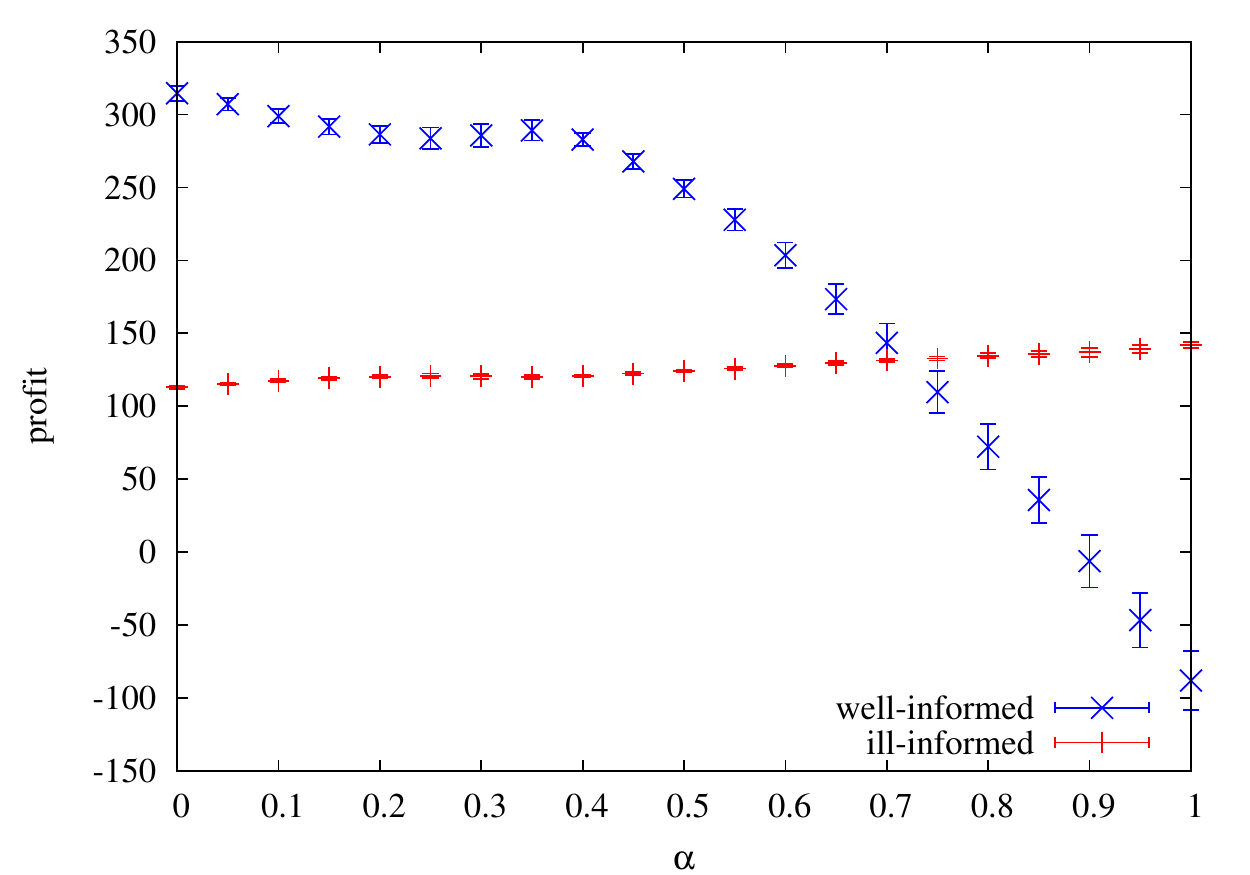}

	\item Fraction of \ills\ $f_{\ii}=0.9$

	\leftFig{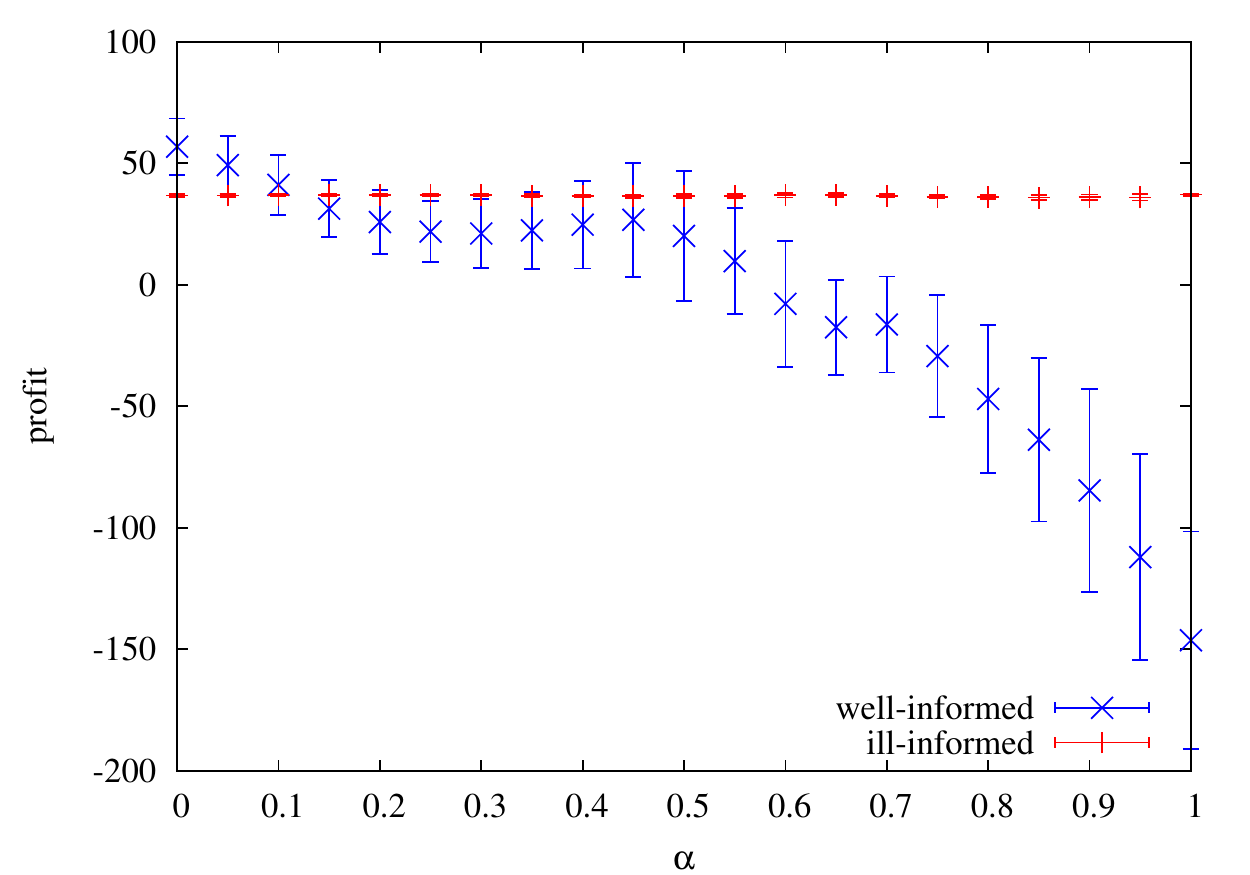}
	\rightFig{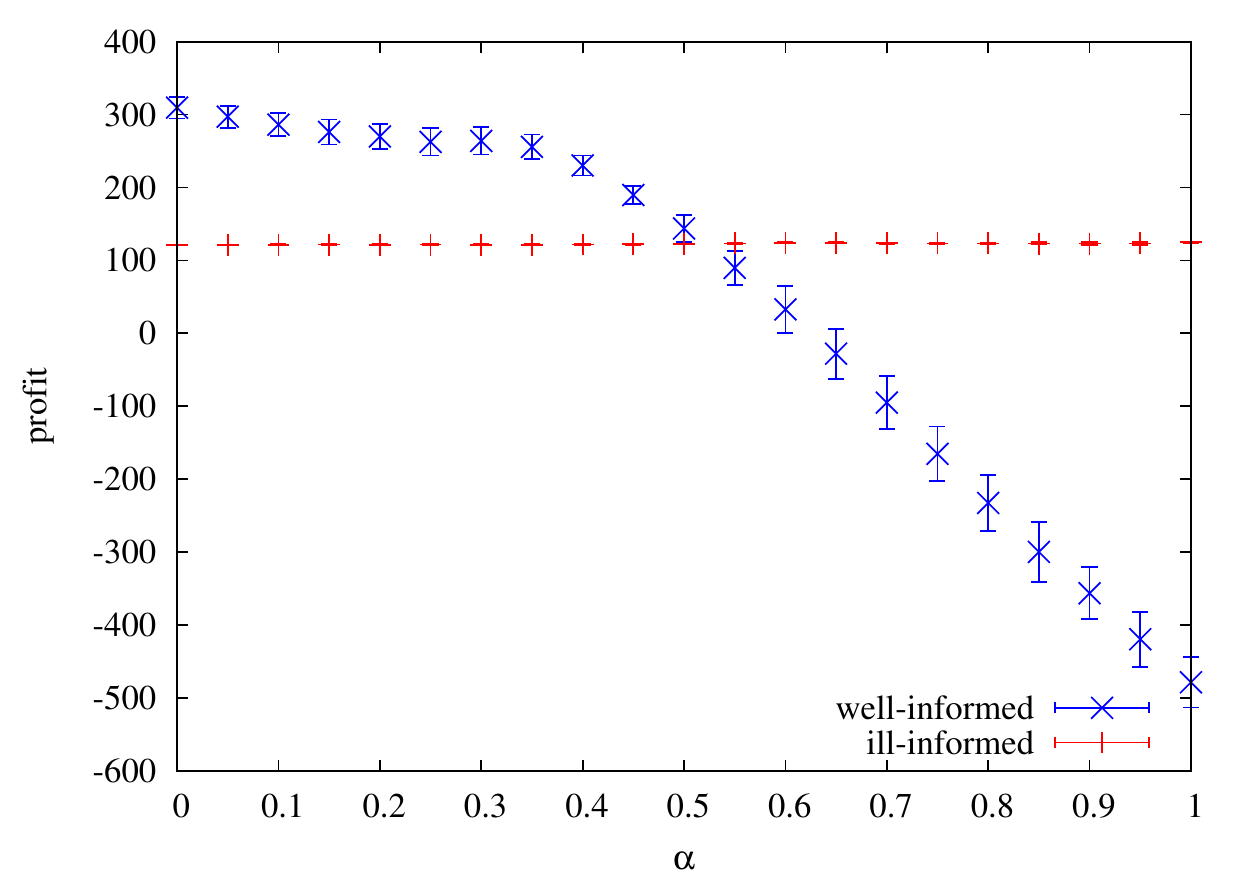}

 \end{itemize}

\end{itemize}
\pagebreak

\subsection{Analysis}
\label{sec:analysis}

First of all, it is evident that in some scenarios \ills\ have a higher profit than \wells. 
We can conclude that under certain circumstances it is more profitable for a consultant to follow the \ii-strategy than being well-informed. 
Some more detailed observations about the results are described in the following.

\subsubsection{Using R1}
One can see that if reputation is used as the only criterion for selection ($\alpha=0$), \wells\ are always better off. 
As the price of the consultants gets more important, the profit of \ills\ generally increases, whereas the profit of \wells\ decreases. 
In most settings the \wells' profit decreases so much that at some point it falls below the profit of the \ills. 
Besides, \ills\ seem to generally benefit from a low profit margin ($\delta$).

An increasing fraction of \ills\ causes the curves of both \ills\ and \wells\ to have a lower slope value (have a look at the values on the y-axis to verify that). 
As a result, the point where the two curves intersect moves to the left, i.e. the \ii-strategy becomes more effective.

A varying $\Delta N_\text{arg}$ impacts also the slope of the curves, as well as their distance. 
Consider for instance the variation of $\Delta N_\text{arg}$ for otherwise fixed parameters: R1, $f_\ii=0.5$ and $\delta=0.5$ (figures on the right side). For $\Delta N_\text{arg}=2$, the \ii-strategy is the better strategy for $\alpha\geq 0.45$; for a $\Delta N_\text{arg}$ of $3$ or $4$, the \wi-strategy is always better -- for $3$ very distinct, and less distinct for $4$.

%An increasing $\Delta N_\text{arg}$ causes the curve of the \wells\ to have a lower slope value (have a look at the values on the y-axis to verify that). 
%This means that the importance of $\alpha$ increases with an increasing $\Delta N_\text{arg}$.

\subsubsection{Using R2}
The same observations as for R1 can also be made for R2. 
The main difference lies in the shape of the curves. 
In particular, the shape of the lines depends on the parity of $\Delta N_\text{arg}$. 
This can be explained as follows. 
If an even number of new arguments gets available each round, \wells\ will always advice even arguments; however, if $Delta N_\text{arg}$ is odd, they switch each round between advising even and odd arguments. 
Since clients retrospectively judge consultants in R2, they will find in the odd case that \wells\ contradict each other over time, which will never be the case in the even case. 
So, in the even case, clients will decrease the reputation of a \well\ only if their most recent consultant was ill-informed and consulted an odd argument. 
This explains also why the profit of \wells\ is much higher and the profit of \ills\ is much lower in the even cases (also in comparison to R1).

Apart from these specific shapes, in the same way as for R1 we can observe that the \ii-strategy yields in many settings higher profits.

\section{Conclusion \& Future Work}
\label{sec:conclusion}

% CONCLUSION
Our simulations suggest that at least four factors have an impact on the profitability of being ill-informed:
\begin{itemize}
	\item the clients' balancing act of choosing a consultant according to his reputation or his price ($\alpha$),
	\item the fraction of ill-informed consultants in the set of consultants ($f_\ii$), 
	\item the speed with which new information enters the system and becomes available to consultants ($\Delta N_\text{arg}$), and 
	\item the profit margin of the consultants ($\delta$).
\end{itemize}
In our simulations, a high $\alpha$ and a high $f_\ii$ made it generally less profitable for consultants to be well-informed.
The observation that an increasing fractions of \ills\ increases their payoff, raises the problem of consultants that follow the crowd, even if this means to stay ill-informed.

% FUTURE WORK
The simulated model in this work is certainly simplifying the complexity of reality. 
There are many ways for making it more realistic. 
For instance, the price computation given by eq. \ref{price_model} is purely reactive; it does not exploit the fact that a lower prices would attract more clients and so could be used in a proactive manner. 
Still, since all consultants computed their price in the same way, it is not clear whether a more complex approach would make the \ii-strategy less profitable.
This is subject to future research.
Also, our consultants pursued two basic strategies (acquiring all arguments becoming available, and acquiring only as needed); this can be extended to more sophisticated strategies, e.g. to be a \well\ for a certain time to boost the reputation, and then become \ill.
Furthermore, the impact of publicity campaigns is completely ignored and would be an interesting continuation of our research.

%\bibliographystyle{abbrv}

%\bibliography{bib,ASPICbibliography}
\end{document}